\documentclass[10pt,twocolumn,letterpaper]{article}

\usepackage{iccv}              % To produce the CAMERA-READY version

\usepackage{makecell}

\definecolor{iccvblue}{rgb}{0.21,0.49,0.74}
\usepackage[pagebackref,breaklinks,colorlinks,allcolors=iccvblue]{hyperref}
\usepackage{listings}

\def\paperID{AIM-41} % *** Enter the Paper ID here
\def\confName{ICCV}
\def\confYear{2025}

\title{Optuna vs Code Llama: \\ Are LLMs a New Paradigm for Hyperparameter Tuning?}

\author{
Roman Kochnev \quad Arash Torabi Goodarzi \quad Zofia Antonina Bentyn \\
Dmitry Ignatov \quad Radu Timofte \\
Computer Vision Lab, CAIDAS \& IFI, University of W\"urzburg, Germany
}
\begin{document}
\maketitle
\begin{abstract}

Optimal hyperparameter selection is critical for maximizing the performance of neural networks in computer vision, particularly as architectures become more complex. This work explores the use of large language models (LLMs) for hyperparameter optimization by fine-tuning a parameter-efficient version of Code Llama using LoRA. The resulting model produces accurate and computationally efficient hyperparameter recommendations across a wide range of vision architectures. Unlike traditional methods such as Optuna, which rely on resource-intensive trial-and-error procedures, our approach achieves competitive or superior Root Mean Square Error (RMSE) while substantially reducing computational overhead. Importantly, the models evaluated span image-centric tasks such as classification, detection, and segmentation, fundamental components in many image manipulation pipelines including enhancement, restoration, and style transfer. Our results demonstrate that LLM-based optimization not only rivals established Bayesian methods like Tree-structured Parzen Estimators (TPE), but also accelerates tuning for real-world applications requiring perceptual quality and low-latency processing. All generated configurations are publicly available in the LEMUR Neural Network Dataset (\url{https://github.com/ABrain-One/nn-dataset}), which serves as an open source benchmark for hyperparameter optimization research and provides a practical resource to improve training efficiency in image manipulation systems.

\end{abstract}
\section{Introduction}
\label{sec:intro}

The choice of training hyperparameters plays a crucial role in determining the learning efficiency and performance of machine learning models. Poorly selected hyperparameters can lead to suboptimal learning and degraded accuracy, while well-chosen configurations can significantly boost performance using the same model architecture.

Large Language Models (LLMs)~\cite{brown2020languagemodelsfewshotlearners} have recently demonstrated the ability to capture complex patterns across various tasks. As their scale and sophistication continue to grow, they are increasingly replacing traditional methods in numerous domains \cite{Gado2025llm, Rupani2025llm, ABrain.NNGPT}. With fine-tuning techniques, LLMs can acquire new capabilities and solve previously unseen problems. In this work, we explore fine-tuning LLMs for hyperparameter optimization (HPO)~\cite{Feurer2019}, positioning them as a novel alternative to traditional methods such as Bayesian Optimization (BO)~\cite{brochu2010tutorialbayesianoptimizationexpensive} and Evolutionary Strategies (ES)~\cite{hansen2001completely}.

Our approach leverages a fine-tuned version of Code Llama~\cite{codellama} to recommend hyperparameters for deep learning models in a one-shot manner. We focus on training hyperparameters for well-known computer vision architectures in the context of image classification, aiming to evaluate whether LLMs can match or surpass the performance of traditional optimization frameworks such as Optuna \cite{10.1145/3292500.3330701}.

Importantly, many of the evaluated architectures, such as ResNet~\cite{he2015deepresiduallearningimage}, MobileNet~\cite{howard2017mobilenetsefficientconvolutionalneural}, and SwinTransformer~\cite{liu2021swintransformerhierarchicalvision}, are widely used as backbones in image manipulation pipelines, including tasks like super-resolution, deblurring, and perceptual enhancement. Efficient hyperparameter tuning of these models has a direct impact on the performance and usability of image restoration systems, particularly in resource-constrained or real-time scenarios. Thus, our findings contribute not only to HPO research, but also to the broader image manipulation community.

%-------------------------------------------------------------------------
\subsection{Related work}
\label{sec:related-work}

Early model-free approaches to the HPO task, such as grid and random search, perform well in low-dimensional hyperparameter spaces. However, as the number of hyperparameters increases, the computational cost of applying these methods also increases. For further information on HPO, we refer to \cite{Feurer2019, 10.5555/3322706.3361994, pr11020349, 10.5555/2188385.2188395, doi:10.1080/1206212X.2021.1974663, https://doi.org/10.1002/widm.1484}.

More advanced techniques for HPO include evolutionary strategies (ES) and Bayesian optimization (BO). Although ES outperforms previous methods in more complex search spaces, its application can still be computationally expensive \cite{https://doi.org/10.1002/widm.1484, 8297018}. 

BO has proven to be an efficient alternative, especially in high-dimensional hyperparameter spaces, and typically exceeds ES in terms of computational efficiency. It operates by constructing a surrogate model, updating its posterior distribution based on observed data, and selecting the next candidate using an acquisition function. The effectiveness of BO strongly depends on the surrogate model used, with the tree-structured Parzen Estimator (TPE) being the focus of this study \cite{10.5555/2986459.2986743}. TPE models the objective function by creating probability distributions for both favorable and unfavorable hyperparameter values. These distributions are iteratively refined during optimization, guiding the search toward regions with a higher likelihood of success. For a more detailed discussion of BO, we refer to \cite{7352306,Feurer2019,10.5555/2999325.2999464,pmlr-v28-bergstra13,pmlr-v80-falkner18a,pmlr-v54-klein17a}.

Applying this framework, Optuna is a flexible hyperparameter optimization tool that supports various surrogate models within the BO paradigm. For this study, we compare Optuna’s TPE-based approach -- due to its effectiveness in navigating complex hyperparameter spaces -- with the Python version of Code Llama (7B and 13B parameter versions).

The Transformer architecture \cite{10.5555/3295222.3295349, devlin-etal-2019-bert}, which underlies most LLMs, has revolutionized a wide range of fields, including natural language processing \cite{NEURIPS2020_1457c0d6,yadav-2022,iu-2023}, code generation \cite{roziere-2023,nijkamp2023codegen,10.1145/3520312.3534862,feng-etal-2020-codebert,10.1145/3650105.3652299}, computer vision \cite{tan-bansal-2019-lxmert,10205465,NEURIPS2022_960a172b}, and complex reasoning tasks \cite{imani-etal-2023-mathprompter,ac1f09077393404a8bea5141d8710259,10.5555/3600270.3601883,10.1007/978-3-031-40292-0_1,hao-etal-2023-reasoning,10.5555/3666122.3666639,10.5555/3600270.3602070}, among many others \cite{20230919,ZHANG2024143,10435998}.

Extensive research has been conducted to evaluate whether LLMs can outperform established HPO methods. For instance, \cite{zhang-2023} demonstrated that LLMs can achieve performance comparable to or better than conventional methods when applied to models such as logistic regression, support vector machines, random forests, and neural networks, even when those methods are paired with various optimization strategies (e.g., \cite{Aboudeshish2025augmentation}). Their study compared various HPO algorithms, including BO with Gaussian process and random forest surrogates, to a random search baseline. \cite{10.1145/3638530.3664163} investigated using LLMs to tune the step-size of the (1+1)-ES algorithm, showing that LLM-driven strategies can effectively compete with traditional methods. \cite{liu-2024} introduced AgentHPO, an iterative LLM-powered HPO tool, which frequently outperforms human-crafted solutions while maintaining interpretability. \cite{liu2024large} proposed LLAMBO, a method that integrates LLMs with BO, demonstrating effective zero-shot warm-starting by leveraging LLMs’ domain knowledge. Lastly, \cite{Mahammadli2024} presented Sequential Large Language Model-Based Optimization (SLLMBO), showcasing how LLMs can outperform traditional BO methods in several scenarios.

The distinguishing feature of our study is the application of fine-tuning to the LLM for the HPO task. Unlike previous work such as \cite{Mahammadli2024}, we focus on locally deployable models. Although we acknowledge that our setup has resource limitations, to the best of our knowledge, the findings and methodology presented here are novel and have not been previously explored.

Unlike most prior LLM-based HPO studies, our work explores the integration of such methods directly into computer vision pipelines relevant for image restoration and enhancement. We emphasize compatibility with tasks commonly found in AIM workflows, such as super-resolution and denoising, where rapid tuning of hyperparameters can substantially impact perceptual quality.
\section{Methodology}

\subsection{Problem Formulation}

The hyperparameter tuning aims to find an optimal configuration \(\lambda^*\) within the search space \(\Lambda\) that minimizes the loss:
\begin{equation}
    \lambda^* = \arg \min_{\lambda \in \Lambda} \mathcal{L}\bigl(f(x; \lambda), y\bigr),
\end{equation}
where \( f(x; \lambda) \) is a neural network parameterized by hyperparameters \(\lambda\), \(x\) represents input data, and \(y\) denotes labels. Traditional methods like Bayesian Optimization (BO) and Evolutionary Strategies (ES) iteratively explore \(\Lambda\) through multiple function evaluations:
\begin{equation}
    \mathcal{L}\bigl(f(x; \lambda_i), y\bigr), \quad \forall i \in \{1, \dots, N\}.
\end{equation}
Although effective, these iterative approaches incur high computational costs, especially for deep learning models.

\subsection{Proposed Solution}

To reduce computational overhead, we fine-tune an LLM to predict optimal hyperparameters in a single inference step. Given a dataset of previously tested hyperparameters:
\begin{equation}
    D = \{(\lambda_i, \mathcal{L}_i)\}_{i=1}^{N},
\end{equation}
where \(\lambda_i\) is a candidate hyperparameter set and \(\mathcal{L}_i\) is its performance metric, the fine-tuned LLM learns to approximate:
\begin{equation}
    \hat{\lambda} = \mathcal{M}(\text{LLM}, D, M),
\end{equation}
with \(\mathcal{M}\) representing the fine-tuned LLM, \(D\) the tuning dataset, and \(M\) the model architecture. Unlike BO or ES, which require sequential evaluations, our approach uses historical tuning data to provide one-shot predictions, significantly reducing the number of costly evaluations.

\subsection{Fine-Tuning Process}

We enhance Code Llama’s \cite{roziere-2023} hyperparameter prediction capability through LoRA \cite{lora} fine-tuning, which modifies only a subset of parameters to minimize overhead. The optimization objective is the following:
\begin{equation}
    \mathcal{L}_{\text{LLM}} = \sum_{i=1}^{N} \left\| \lambda_i - \mathcal{M}(\text{LLM}, D, M) \right\|^2.
\end{equation}
Code Llama is initialized with pre-trained weights and further fine-tuned on a curated dataset of hyperparameter-performance pairs, enabling it to generalize across various architectures.

\subsection{Evaluation Strategy}

We compare our fine-tuned LLM with Optuna by measuring the RMSE of accuracy-based errors. Specifically, for each trial \(i\), we define 
\(\epsilon_i = 1 - a_i\), 
where \(a_i\) is the observed accuracy. The RMSE is then computed as
\(\mathrm{RMSE} = \sqrt{\frac{1}{N}\sum_{i=1}^{N} (\epsilon_i)^2})\),
providing a measure of how closely the trials match ideal (error-free) performance. In addition, we assess stability across trials and analyze the impact of fine-tuning cycles. The goal is to demonstrate that Code Llama provides competitive hyperparameter recommendations with reduced computational cost.
\section{Implementation}
\label{sec:implementation}

\begin{figure*}
    \centering
    \includegraphics[width=0.75\linewidth]{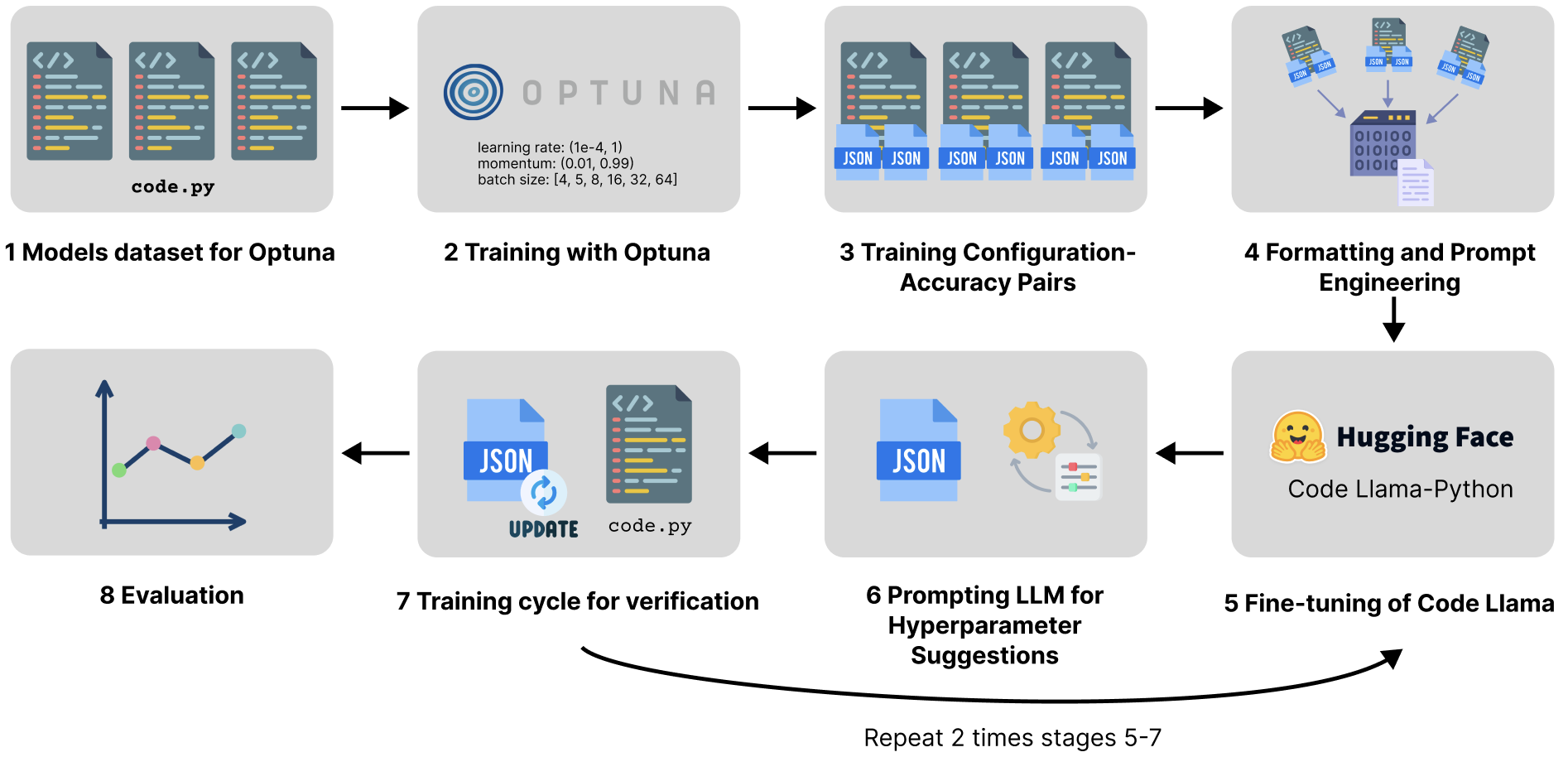}
    \caption{Generation of training hyperparameters for our neural network dataset with Optuna and Code Llama.}
    
    \label{fig:pipeline-overview}
\end{figure*}

In this section, we elaborate on the process we implemented to obtain our results. This process consisted mainly of eight main parts depicted in Figure \ref{fig:pipeline-overview}. Each of these parts is explained in its respective subsection below.

\subsection{Dataset Preparation and Initial Hyperparameter Tuning} 
\label{ssc:dataset-prep}

For our experiments, we standardized the implementation of computer vision models available in the TorchVision software package \cite{torchvision} and evaluated their performance on the CIFAR-10 dataset \cite{cifar-10} using various hyperparameter configurations.

Additionally, we prepared the source code for the following neural network models: RNN~\cite{elman1990finding}, Long Short-Term Memory (LSTM)~\cite{hochreiter1997long}, and Llama 3~\cite{grattafiori2024llama3herdmodels}, targeting text generation tasks. These models were trained on the Salesforce/WikiText dataset~\cite{merity2016pointersentinelmixturemodels} taken from the HuggingFace platform.

To determine the optimal hyperparameters of each model, we used the Optuna framework~\cite{10.1145/3292500.3330701}. 
The hyperparameters being tuned included the learning rate, batch size, and momentum, each sampled from specific ranges or sets of values. The learning rate was chosen from a continuous range between 0.0001 and 1, while the batch size was selected from the set $\{4, 5, 8, 16, 32, 64\}$. The momentum, which influences the convergence speed of the training process, was selected from a range of 0.01 to 0.99.
Each model was trained using these hyperparameter configurations over different numbers of epochs, namely 1, 2, and 5 (for some models) for each configuration.
After completion of each training run, the selected hyperparameter values and the accuracy achieved were recorded. The results, including the final accuracy for each set of hyperparameters, were then saved in a JSON file, providing a structured dataset of hyperparameter configurations and their corresponding performance metrics. This data collection process enabled a thorough analysis of how different hyperparameter settings influenced the performance of both image classification and text generation models.

This dataset preparation resulted in 3700 entries including all the hyperparameter-accuracy pairs determined by Optuna for each of the 17 models mentioned. The resulting dataset was released publicly as part of the LEMUR NN Dataset~\cite{lemurNNpaper}, which serves as an open source benchmark for research in hyperparameter optimization and AutoML.

%%%%%%%% Replaced under anonymous submission conditions %%%%%%%%

\subsection{Code Llama And Fine-Tuning}

For this project, the Code-Llama-Python~\cite{codellama} model was chosen as the base model for fine-tuning to recommend optimal hyperparameters for neural network architectures. 
This decision was guided by several key factors, including open-source availability, performance, and the fact that this version of Code Llama is predominantly trained on Python code.

To generate valid and meaningful hyperparameter suggestions using Code-Llama-Python, we fine-tuned the base model starting with its pre-trained checkpoint available on HuggingFace. 
For fine-tuning, we employed the LoRA technique \cite{lora}, using a rank of 32, an alpha set to 16, and a dropout rate of 0.05. The training was conducted over 35 epochs.

\begin{lstlisting}[caption={Prompt format for fine-tuning}, captionpos=b, label={listing:prompt_format}, basicstyle=\normalsize]
    <System Prompt>
    
    ### Input:
    <Instruction including 
     code and wanted accuracy>
    
    ### Response:
    <Response including 
     the hyperparameters 
     that would achieve 
     the wanted accuracy>
\end{lstlisting}

Using our dataset we created, as explained in Section \ref{dataset-prep}, a prompt format to use for the fine-tuning including a system prompt telling the model its role as a hyperparameter suggestion tool, an instruction asking for the three hyperparameters we are tuning for a given model mentioning its implementation code to achieve a mentioned accuracy and a response which includes the hyperparameters that would achieve the named accuracy in the instruction.
The prompt format we chose is the following common choice for fine-tuning shown in Listing \ref{listing:prompt_format}.

After each fine-tuning cycle, the LLM’s performance is evaluated by prompting it to suggest the best possible training hyperparameters for the given models.
The existing dataset is expanded by adding hyperparameter values generated by the Code-Llama-Python model. After the generation of these values, each set of hyperparameters undergoes validation by running a training process for the specific model at hand.
The hyperparameters generated, consisting of learning rate, momentum, and batch size, are applied to the training configuration. The resulting model accuracy is then compared with the initially desired accuracy specified in the dataset. This comparison serves as a feedback loop, where the accuracy obtained is used to refine and further train the Code-Llama-Python model, improving its hyperparameter prediction capabilities for subsequent iterations. This process ensures that the model evolves to produce increasingly
optimized hyperparameter sets, leading to enhanced performance in future predictions.

This technique increased the size of our dataset to 7107 entries in the second cycle, which resulted in the model having a robust statistical representation of training hyperparameters for neural network models.
\section{Evaluation}

% Training of computer vision models and fine-tuning of Code Llama are performed using the Linux docker image on NVIDIA GeForce RTX 3090/4090 GPUs of the CVL Kubernetes cluster.

%%%%%%%% Replaced under anonymous submission conditions %%%%%%%%
Training of computer vision models and fine-tuning of Code Llama are performed using the AI Linux docker image abrainone/ai-linux~\footnote{AI Linux: \url{https://hub.docker.com/r/abrainone/ai-linux}} on NVIDIA GeForce RTX 3090/4090 GPUs of the CVL Kubernetes cluster at the University of W\"urzburg.

\begin{figure*}
    \centering
    \includegraphics[width=0.66\linewidth]{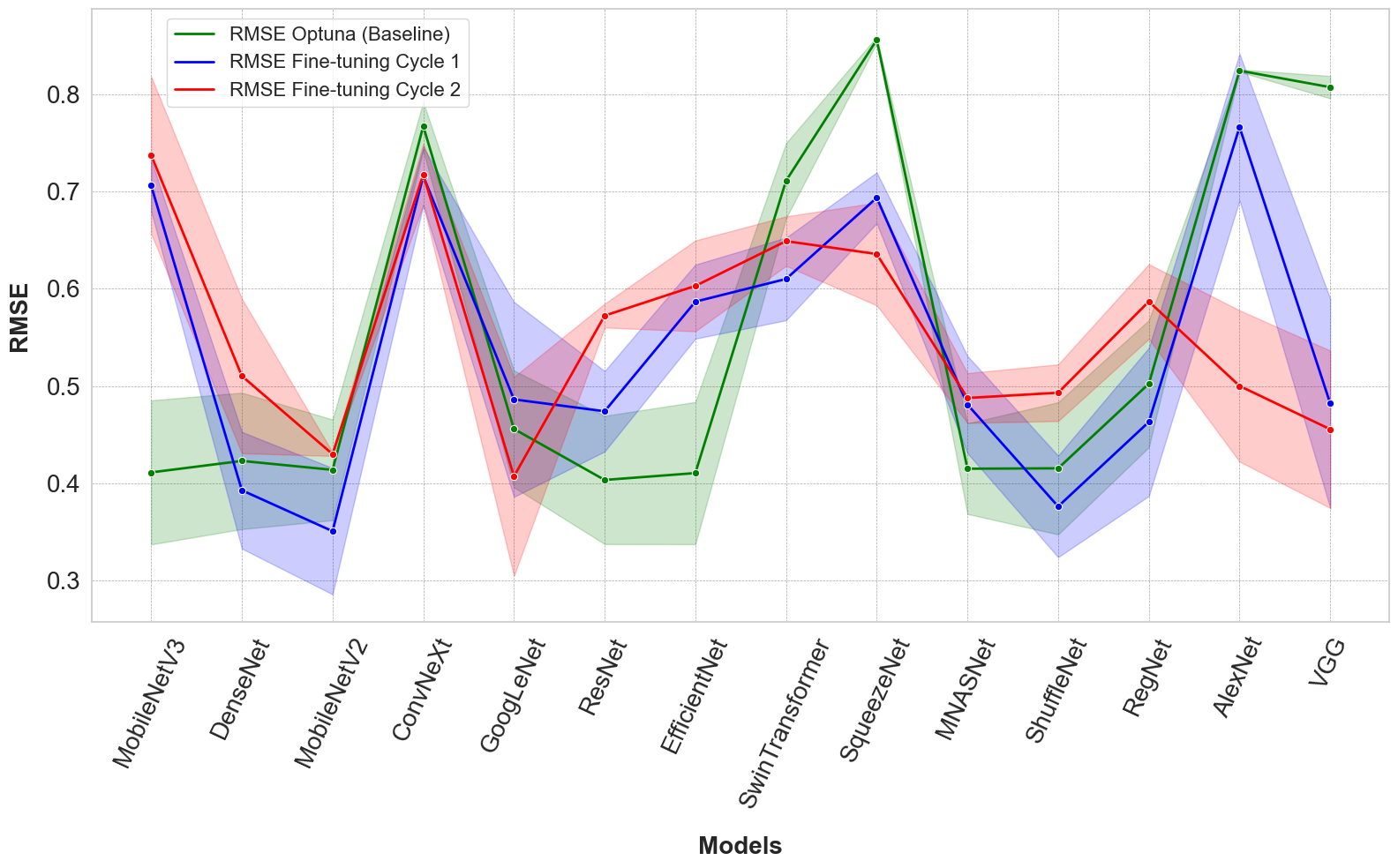}
    \caption{RMSE values for 14 computer vision models, comparing results of Optuna (in green) and obtained after fine-tuning Code Llama (fine-tuned cycle 1 in blue and fine-tuned cycle 2 in red). The shaded area around the line indicates 95\% confidence interval, reflecting the uncertainty of the RMSE estimates. This visualization aids in assessing the performance of the models across different datasets and epochs.}  
    \label{figure:2_Optuna_FTs}
\end{figure*} 

\subsection{Pre-Trained Code Llama}

Upon analyzing the responses from the original pre-trained version of Code Llama, we observed that the model returned a total of 1006 relevant records out of 2980 responses, yielding a relevance rate of approximately 33.77\%. This indicates that approximately one-third of the responses were deemed pertinent to the queries posed. However, a notable observation was the presence of a significant number of answers that lacked numerical values, which could affect the utility of the responses for tasks that require quantitative analysis.

Additionally, we identified certain patterns in the parameters suggested by the pre-trained Code Llama. For instance, the frequently suggested batch size value was 0, raising concerns about its applicability in real-world scenarios. Typically, a batch size of 0 is impractical and can cause confusion among users who attempt to implement the recommendations of the model. Furthermore, the most common learning rate suggested in the responses was 0.01. This learning rate is often a standard starting point in many machine learning contexts, but without fine-tuning, its effectiveness may vary depending on the specific dataset and task at hand.

\subsection{Baseline Results and Training Efficiency}

Although accuracy is essential for our tasks, RMSE was chosen for its ability to capture fine-grained error dynamics. RMSE provides a continuous measure of performance variability across trials and fine-tuning cycles, providing insights into model stability and robustness.

The fine-tuned Code Llama consistently achieves lower RMSE values than the baseline Optuna results for most models used in fine-tuning, demonstrating its effectiveness in hyperparameter optimization. As shown in Figure~\ref{figure:2_Optuna_FTs}, models such as SwinTransformer~\cite{liu2021swintransformerhierarchicalvision}, SqueezeNet~\cite{iandola2016squeezenetalexnetlevelaccuracy50x}, and VGG~\cite{simonyan2015deepconvolutionalnetworkslargescale} experience significant error reductions after fine-tuning, highlighting substantial improvements in hyperparameter recommendations. Furthermore, architectures such as ConvNeXt~\cite{liu2022convnet2020s} and GoogLeNet~\cite{szegedy2014goingdeeperconvolutions} show moderate but meaningful enhancements, further validating the adaptability of Code Llama across different types of networks. The contraction of shaded 95\% confidence intervals for most architectures indicates improved stability and reduced variability in hyperparameter predictions. Since no substantial reductions in RMSE were observed beyond the second fine-tuning cycle, further fine-tuning was deemed unnecessary.

\begin{table}[ht]
\fontsize{8.5}{8.5}\selectfont
\centering
\begin{tabular}{c c c c c }
\toprule
\textbf{Method} & \textbf{Trial} & \textbf{RMSE} & \textbf{\(\sigma\)} & \textbf{95\% Conf. Int.}\\\midrule
Optuna & All & 0.589 & 0.219 & [0.581, 0.597] \\ 
& Best & 0.416 & 0.115 & [0.375, 0.456] \\\midrule
& Fine-tuning 1 & 0.563 & 0.182 & [0.556, 0.570] \\ 
LLM & Fine-tuning 2 & 0.567 & 0.159 & [0.563, 0.572] \\ 
& Best & \textbf{0.404} & 0.118 & [0.358, 0.480] \\ 
& One-shot & \textbf{0.533} & 0.162 & [0.470, 0.596]                   \\\bottomrule
\end{tabular}
\caption{RMSE, standard deviation \((\sigma\)), and 95\% confidence intervals for Optuna: all obtained accuracies (all) and only the best obtained accuracies (best) and Code Llama: fine-tuning cycle 1 (fine-tuning 1), fine-tuning cycle 2 (fine-tuning 2), the best obtained accuracies (best) and one-shot prediction.}
\vspace{-4mm}
\label{table:Average_RMSE_All}
\end{table}

The results in Table \ref{table:Average_RMSE_All} show that fine-tuning using Code Llama resulted in an improvement in RMSE compared to the Optuna baseline. After the first round of fine-tuning, the RMSE decreased from 0.589 to 0.563, and the confidence interval narrowed, indicating increased stability. The second round of fine-tuning was able to maintain these improvements with minor changes, while the standard deviation also decreased, confirming the increased stability of the model. In this context, the “one-shot” prediction refers to the initial result obtained from the fine-tuned Code Llama after completing its first fine-tuning cycle for a specific model. The largest reduction in RMSE was observed in the one-shot mode (to 0.533), demonstrating the potential of this approach. More details on the one-shot approach can be found in Section \ref{subsection:Prediction_Dynamics_Across_Epochs}.

\begin{figure*}[h!]
    \centering
    \setlength{\tabcolsep}{2pt}
    
    \begin{subfigure}{0.32\textwidth}
        \centering
        \includegraphics[width=\linewidth]{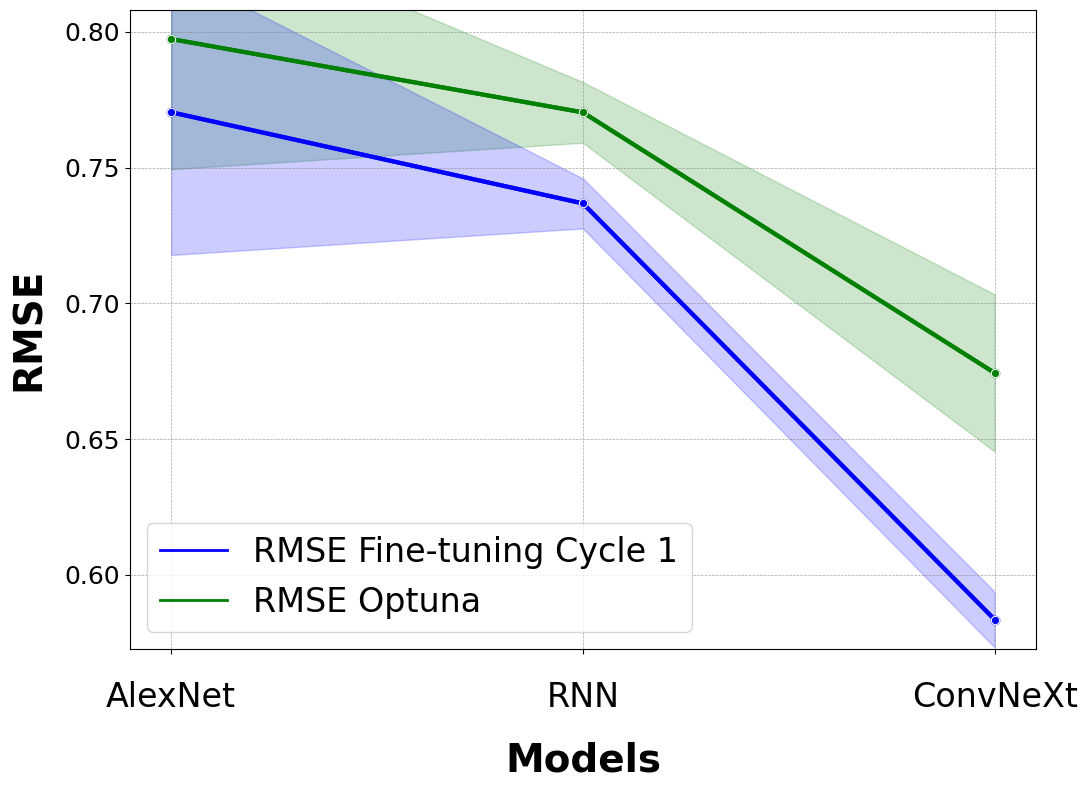}
        \caption{Models included in fine-tuning along with their 5-epoch versions}
        \label{figure:8_9_epochs_used_in_FT_5}
    \end{subfigure}
    \hfill
    \begin{subfigure}{0.32\textwidth}
        \centering
        \includegraphics[width=\linewidth]{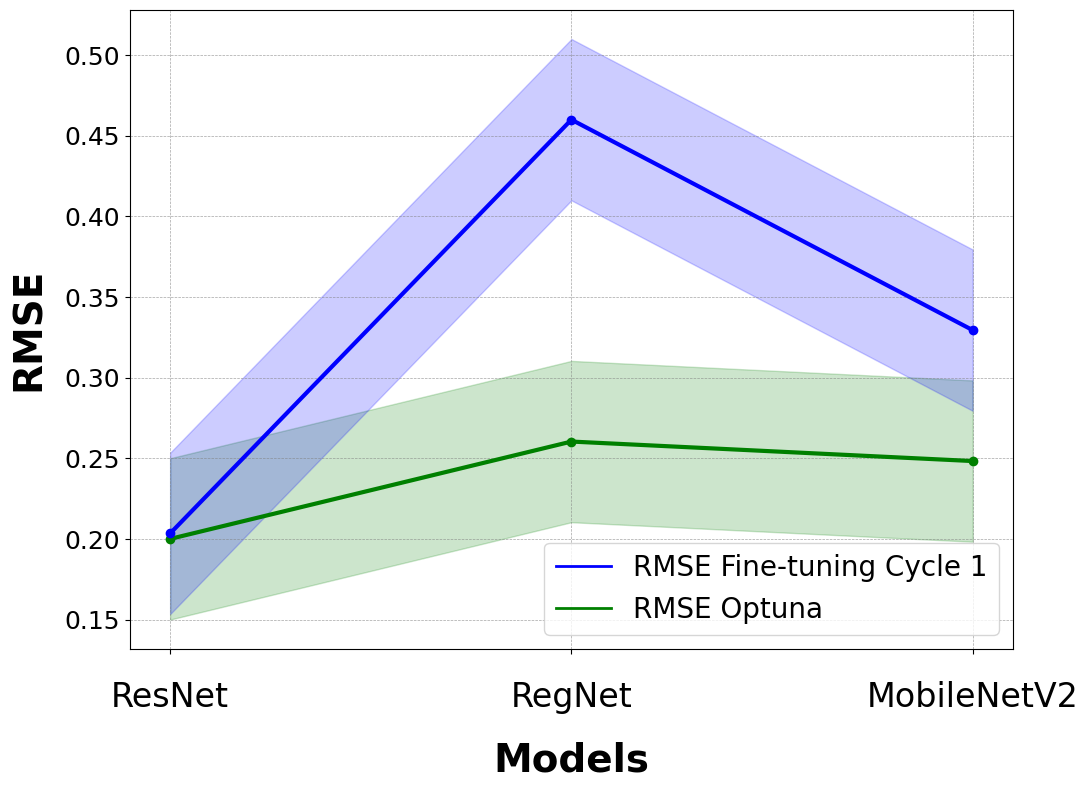}
        \caption{Models used in fine-tuning, but their 5-epoch versions were excluded}
        \label{figure:8_9_epochs_not_used_in_FT_5}
    \end{subfigure}
    \hfill
    \begin{subfigure}{0.32\textwidth}
        \centering
        \includegraphics[width=\linewidth]{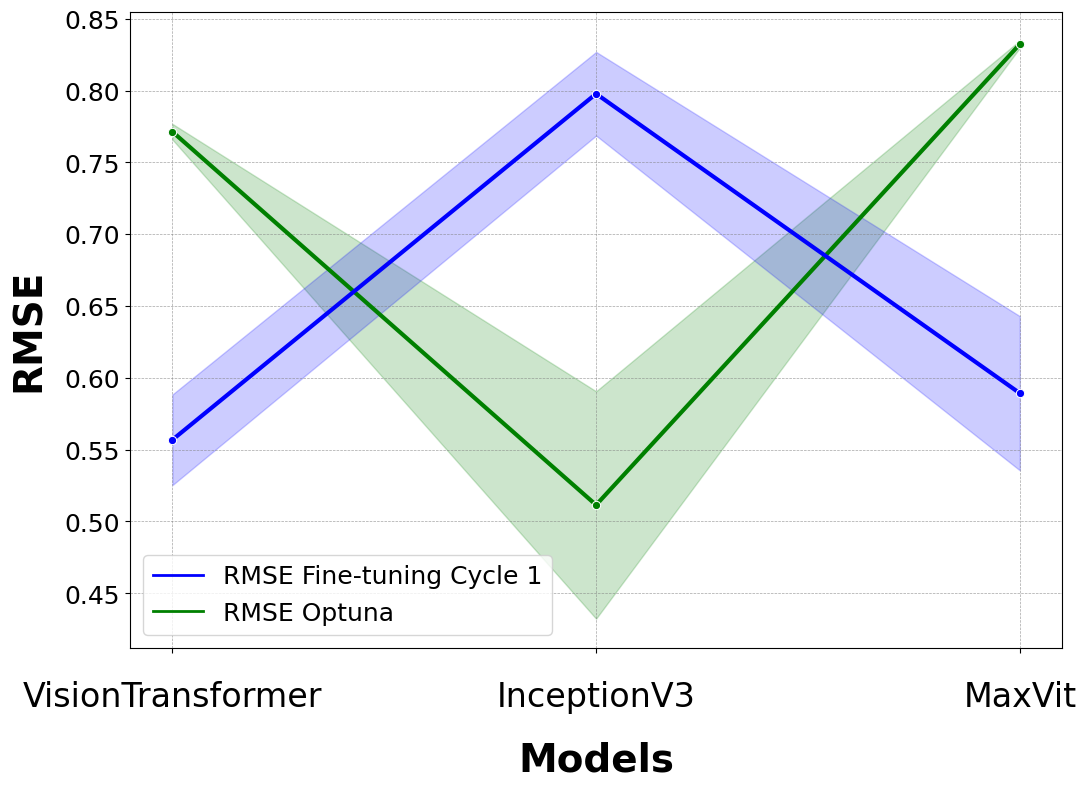}
        \caption{Models with 1 and 2 training epochs taken from the test dataset}
        \label{figure:6_Generalization_3_models}
    \end{subfigure}

    \caption{RMSE values of computer vision models after 5 epochs. The left figure presents models that were used in fine-tuning, along with their 5-epoch versions. The middle figure shows models that were part of the fine-tuning process, but their 5-epoch versions were excluded. The right figure illustrates RMSE values for models with only 1 and 2 epochs, which were part of the test dataset and were never used during fine-tuning. The shaded areas indicate 95\% confidence intervals.}
    \label{fig:rmse_epochs}
\end{figure*}

The 95\% confidence intervals in this study were calculated using Student’s t-distribution. Specifically, the standard deviation (\(\sigma\)) of the errors was calculated for each trial and the standard error (\(SE = \sigma / \sqrt{n}\), where \(n\) is the number of trials) was used to determine the interval bounds. The final confidence intervals were obtained as \( RMSE \pm t_{\alpha/2, n-1} \times SE \), where \( t_{\alpha/2, n-1} \) is the critical value from the t-distribution. This method provides a statistically rigorous estimate of the uncertainty around the reported RMSE values. For more information on confidence intervals in statistical context, refer to \cite{statistics}.

Additionally, focusing on the best accuracy results for each method, we see that Optuna best accuracy (Optuna Best) and Code Llama best accuracy (LLM Best) achieved lower RMSE values of 0.416 and 0.404, respectively, outperforming the overall RMSE for each approach. The LLM Best RMSE of 0.404, accompanied by a slightly higher standard deviation of 0.118 compared to Optuna Best’s 0.115, nonetheless demonstrates Code Llama’s strong performance in reaching optimal parameter configurations. The confidence interval for LLM Best ([0.358, 0.480]) also overlaps closely with Optuna Best ([0.375, 0.456]), showing comparable stability in the highest-performing configurations of each method. These findings underscore the capability of Code Llama not only to improve model stability with fine-tuning but also to achieve highly competitive accuracy levels with optimal parameter selection.

\subsection{Model Performance Analysis}

In our experiment, we used models from both the training and test datasets to demonstrate how fine-tuning with Code Llama handles different scenarios. First, we tested three models that were retrained for 1, 2, and 5 epochs to evaluate the impact of fine-tuning on the models used in training. Then, we considered three models that were not retrained for the same 5 epochs to evaluate the generalization ability of fine-tuning. Finally, we tested three models from the test dataset that were not used in the training at all to see how fine-tuning can handle new, previously unseen models.

{\textbf{Models from train dataset.}}
The comparison between models included in fine-tuning along with their 5-epoch versions (AlexNet~\cite{NIPS2012_c399862d}, RNN~\cite{elman1990finding}, and ConvNeXt~\cite{liu2022convnet2020s}) and models whose 5-epoch versions were excluded from fine-tuning (ResNet~\cite{he2015deepresiduallearningimage}, RegNet~\cite{radosavovic2020designingnetworkdesignspaces}, MobileNetV2~\cite{sandler2019mobilenetv2invertedresidualslinear}) highlights the impact of fine-tuning on RMSE values. As shown in Figure~\ref{figure:8_9_epochs_used_in_FT_5}, models that underwent fine-tuning demonstrate consistently lower RMSE values, with significantly reduced confidence intervals, reflecting improved stability and robustness in hyperparameter recommendations. ConvNeXt, in particular, exhibits a notable decrease in RMSE, confirming the effectiveness of fine-tuning. Models whose 5-epoch versions were not used in fine-tuning (see Figure~\ref{figure:8_9_epochs_not_used_in_FT_5}) exhibit higher RMSE values in the LLM’s predictions compared to Optuna, indicating room for improvement in generalization. However, the fine-tuned model maintains reasonable stability, as reflected in confidence intervals that, while broader, still reflect consistent behavior across trials.

\begin{figure*}[h!]
    \centering
    \setlength{\tabcolsep}{1pt}
    \renewcommand{\arraystretch}{5.0}
    \begin{tabular}{cc}
        
        % Line 1
        \begin{minipage}{0.43\textwidth}
            \centering
            \includegraphics[width=\linewidth]{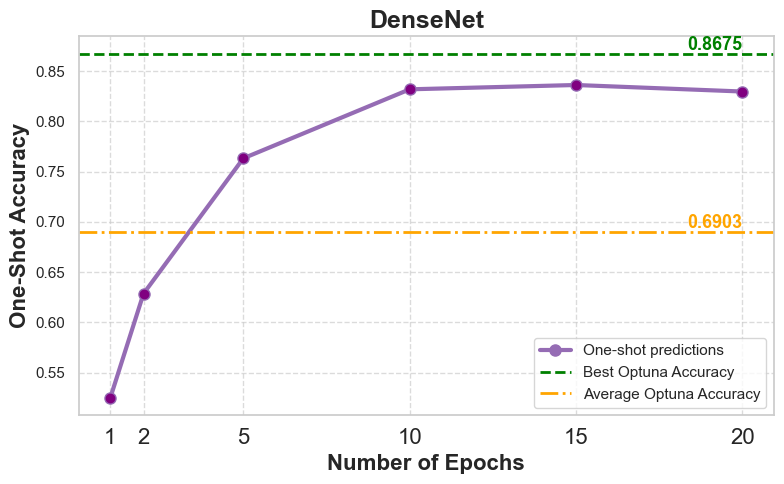}
            % \captionof{figure}{Image 1}
        \end{minipage} &
        \begin{minipage}{0.43\textwidth}
            \centering
            \includegraphics[width=\linewidth]{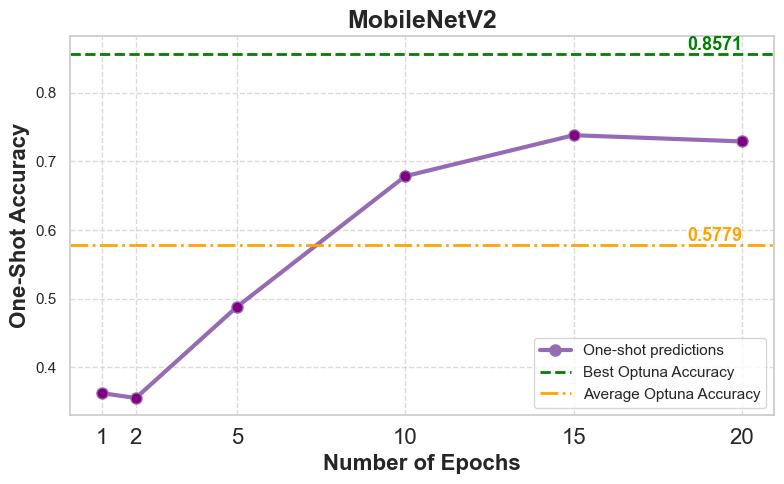}
            % \captionof{figure}{Image 2}
        \end{minipage} \\
        
        % Line 2
        \begin{minipage}{0.43\textwidth}
            \centering
            \includegraphics[width=\linewidth]{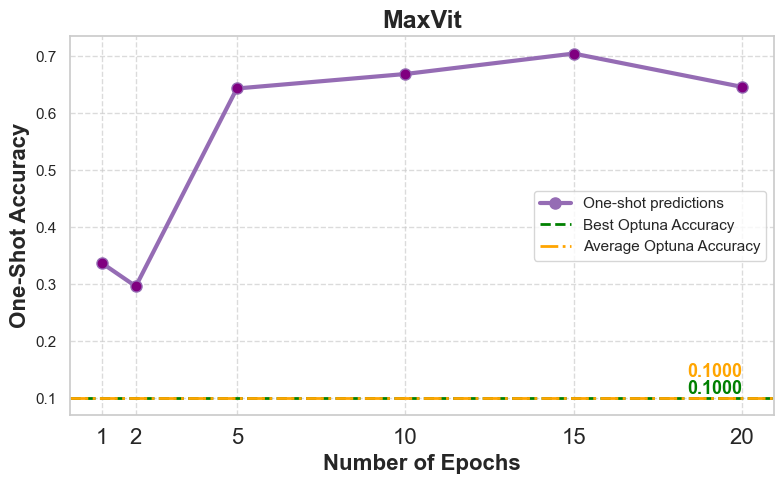}
            % \captionof{figure}{Image 3}
        \end{minipage} &
        \begin{minipage}{0.43\textwidth}
            \centering
            \includegraphics[width=\linewidth]{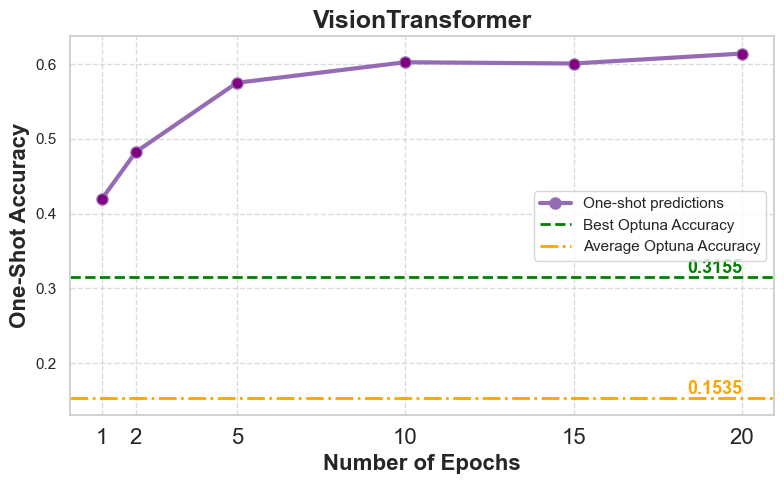}
            % \captionof{figure}{Image 4}
        \end{minipage} \\
    \end{tabular}
\caption{Results of accuracy in one-shot predictions for 4 computer vision models for 1, 2, 5, 10, 15, and 20 epochs, compared with average and best accuracy obtained using Optuna over 10 trials. Top row: training dataset models (DenseNet and MobileNetV2) which participated in fine-tuning with Code Llama for 1 and 2 epochs. Bottom row: testing dataset models (MaxVit and VisionTransformer) which did not participate in fine-tuning, demonstrating Code Llama’s generalization ability.} 
\label{figure:11_4_models_One_Shot_with_Optuna}
\end{figure*}

{\textbf{Models from test dataset.}}
To evaluate the performance of the fine-tuned model, we tested it on three previously unseen architectures: InceptionV3~\cite{szegedy2015rethinkinginceptionarchitecturecomputer}, VisionTransformer~\cite{dosovitskiy2021imageworth16x16words}, and MaxViT~\cite{tu2022maxvitmultiaxisvisiontransformer}. These models were not part of the fine-tuning process, which means that Code Llama had no prior exposure to them. The fine-tuned model generated the corresponding hyperparameters, and we compared the resulting RMSE values against those obtained using the Optuna framework.

As shown in Figure~\ref{figure:6_Generalization_3_models}, the most notable improvement is observed for VisionTransformer and MaxViT, where the fine-tuned Code Llama achieves significantly lower RMSE values than Optuna, indicating that its hyperparameter recommendations lead to more stable and effective optimization for these architectures. Conversely, for InceptionV3, Optuna demonstrates a clear advantage, yielding a lower RMSE, although with wider confidence intervals, suggesting greater variability in its predictions. This contrast highlights that while fine-tuned Code Llama successfully generalizes to certain unseen models, its effectiveness may depend on the architectural similarity to those encountered during training. Nevertheless, its ability to generate competitive hyperparameter recommendations without additional tuning reinforces its potential for efficient and automated hyperparameter selection, particularly in scenarios requiring rapid model deployment.

\subsection{Prediction Dynamics Across Epochs \label{subsection:Prediction_Dynamics_Across_Epochs}}
This analysis examines the progression of model accuracy in one-shot predictions for both training and test sets across multiple epochs (1, 2, 5, 10, 15, and 20).

In this experiment, we analyze hyperparameter recommendations for a diverse set of models. Although a complete evaluation was performed for all models in our study, we present detailed results for four representative architectures: MobileNetV2 and DenseNet~\cite{huang2018denselyconnectedconvolutionalnetworks}, which were part of the training set (participated in Code Llama fine-tuning with data from 1 and 2 epochs), and MaxViT and VisionTransformer, which belong to the test set (not involved in Code Llama fine-tuning). This selection provides a balanced comparison between models seen and unseen during fine-tuning.

The experiment illustrates the accuracy of one-shot predictions across multiple epochs (1, 2, 5, 10, 15, and 20), as obtained using the fine-tuned Code Llama, and compares them to the best and average accuracy values achieved by Optuna. As shown in Figure~\ref{figure:11_4_models_One_Shot_with_Optuna}, training models (MobileNetV2 and DenseNet) show a consistent increase in prediction accuracy as the number of epochs increases. Their performance surpasses the average Optuna accuracy and closely approaches the best Optuna results, confirming the effectiveness of fine-tuning on seen models.

For test models (MaxVit and VisionTransformer), Code Llama also demonstrates strong generalization. Notably, MaxVit achieves a one-shot accuracy close to 0.7, significantly outperforming both the average and the best Optuna results, which remain at 0.1. VisionTransformer also steadily improves with more training epochs, showing superior performance relative to Optuna's average and best scores. Detailed results for additional models can be found in the Supplementary Material (Figures 6–7). These outcomes highlight Code Llama's ability to generate competitive hyperparameter configurations even for unseen architectures.

Although an exhaustive search can theoretically yield optimal hyperparameter values, the associated time costs are a considerable constraint. In real-world model training, exhaustive search becomes highly impractical, underscoring the value of efficient hyperparameter tuning methods such as Code Llama.

\vspace{-1mm}
\subsection{Training Trends and Performance Comparison
\label{subsection:Training_Trends_and_Performance_Comparison}}
\vspace{-3mm}

As part of the experiment, we decided to compare the results of 100 training runs obtained with Optuna with the best values of 100 hyperparameter predictions after fine-tuning with Code Llama, as well as the one-shot prediction accuracy of fine-tuned Code Llama. One-shot metrics were also obtained for each model, providing an indication of how the models can perform when predicted once using the hyperparameters provided by Code Llama.

\begin{figure*}[h!]
    \centering
    \setlength{\tabcolsep}{4pt}
    \renewcommand{\arraystretch}{8.0}
    \begin{tabular}{cc}
        
        % Line 1
        \begin{minipage}{0.42\textwidth}
            \centering
            \includegraphics[width=\linewidth]{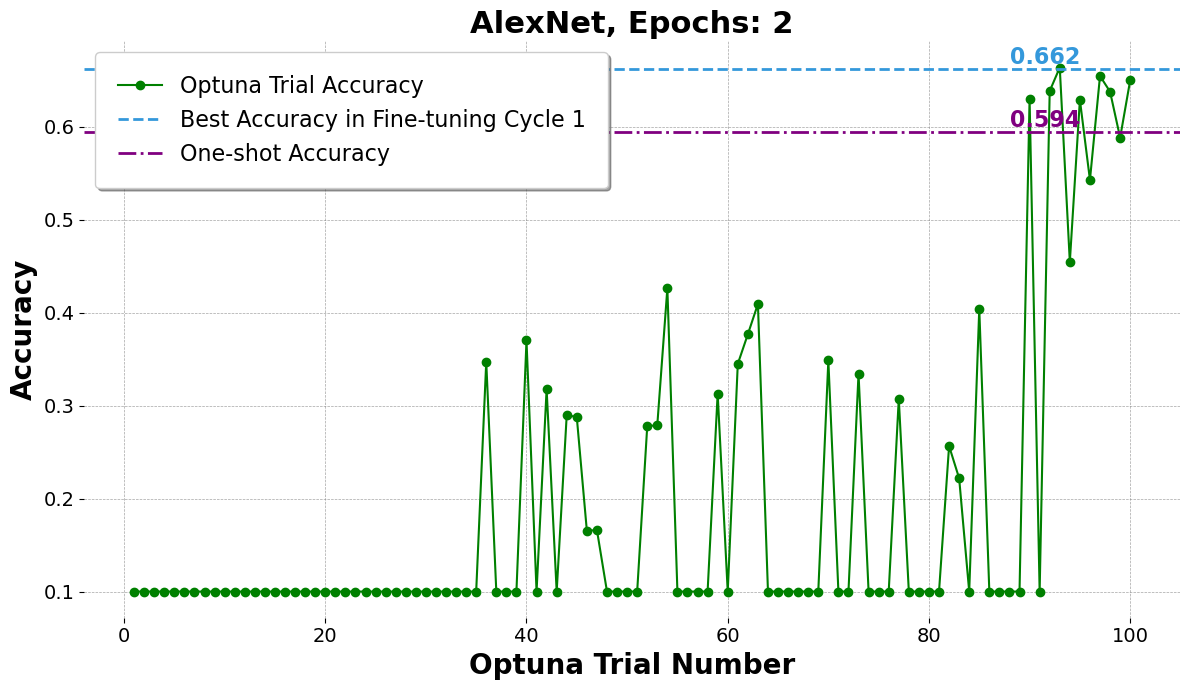}
            % \captionof{figure}{Image 1}
        \end{minipage} &
        \begin{minipage}{0.42\textwidth}
            \centering
            \includegraphics[width=\linewidth]{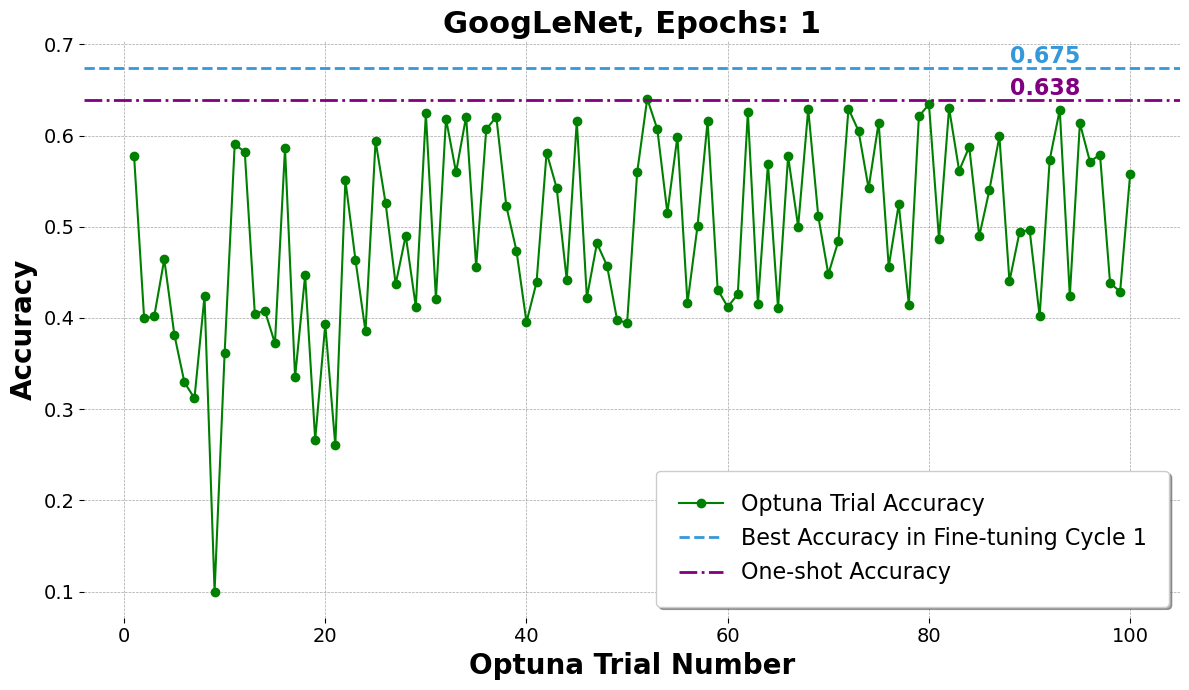}
            % \captionof{figure}{Image 3}
        % Line 2
        \end{minipage} \\
        \begin{minipage}{0.42\textwidth}
            \centering
            \includegraphics[width=\linewidth]{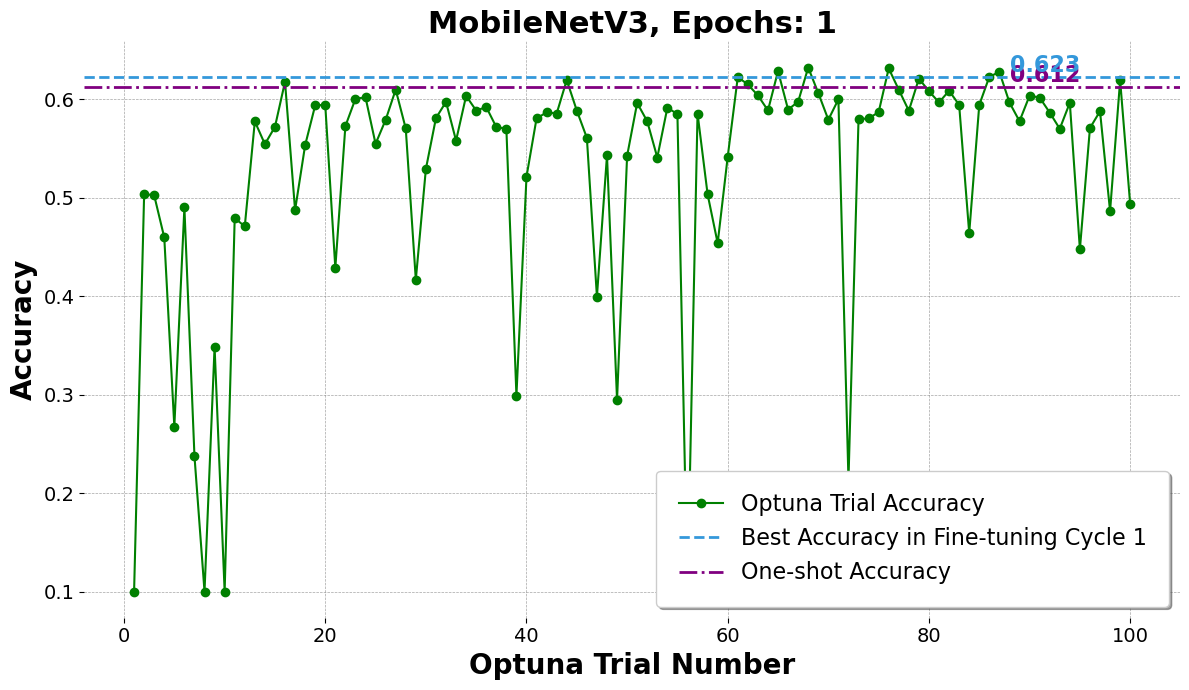}
            % \captionof{figure}{Image 4}
        \end{minipage} &
        \begin{minipage}{0.42\textwidth}
            \centering
            \includegraphics[width=\linewidth]{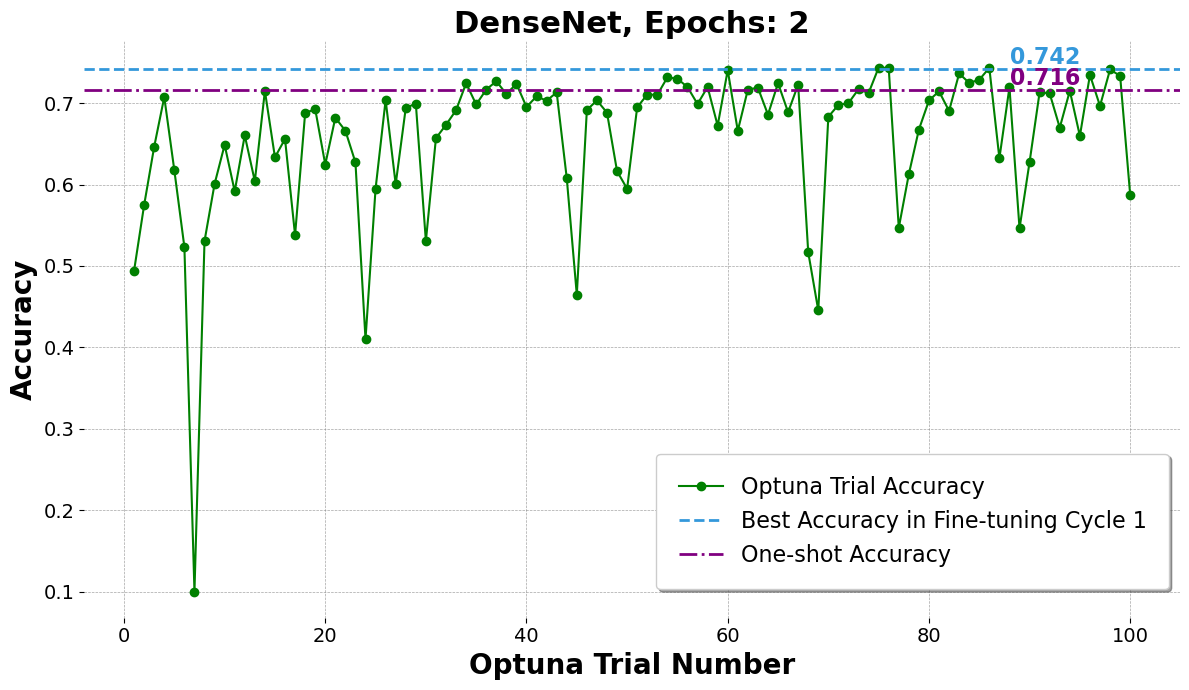}
            % \captionof{figure}{Image 3}
        \end{minipage}
    \end{tabular}
\caption{Comparative analysis of the accuracy of 4 computer vision models using Optuna (green) on 100 trials, the best accuracy value on hyperparameters obtained from fine-tuned Code Llama ( blue) and a one-shot prediction (purple) based on fine-tuned Code Llama.} 
\label{figure:10_6_models_Optuna_and_two_FT_lines}
\vspace{-4.2mm}
\end{figure*}

Analysis of the six plots shown in Figure \ref{figure:10_6_models_Optuna_and_two_FT_lines} shows that fine-tuning significantly improves the accuracy of the models: in each plot, the best results after fine-tuning (blue line) outperform the average results achieved by Optuna (green dots). This demonstrates that fine-tuning allows the models to achieve higher accuracy by optimizing the hyperparameters. Furthermore, the one-shot predictions (purple line) provide a high level of accuracy, approaching the results of a full round of fine-tuning, highlighting the potential of this approach to produce fast and accurate predictions.

Optuna, as seen from the green dots in the graphs, shows a significant spread in accuracy values, especially for a model such as AlexNet, GoogLeNet, and MobileNetV3, indicating high variability in the results. At the same time, fine-tuning shows more stable results, confirming its effectiveness in achieving consistent accuracy. The greatest effect of fine-tuning is seen for DenseNet and MobileNetV3, where the blue line is noticeably higher than the purple and green ones, indicating significant improvements. For some models, like GoogLeNet, the difference between the best values after fine-tuning and the one-shot predictions is less pronounced, indicating limited potential for improvement for individual architectures, but even here fine-tuning maintains a small advantage in accuracy. Detailed results for additional models can be found in the Supplementary Material (Figures 8-10).

%%%%%%%% Replaced under paper length constraints submission conditions %%%%%%%%
% Optuna, as seen from the green dots in the graphs, shows a significant spread in accuracy values, especially for a model like ConvNeXt and MobileNetV3, indicating high variability in results. At the same time, fine-tuning shows more stable results, confirming its effectiveness in achieving consistent accuracy. The greatest effect of fine-tuning is seen for DenseNet and MobileNetV3, where the blue line is noticeably higher than the purple and green ones, indicating significant improvements. For some models, like GoogLeNet, the difference between the best values after fine-tuning and the one-shot predictions is less pronounced, indicating limited potential for improvement for individual architectures, but even here fine-tuning maintains a small advantage in accuracy. Detailed results for additional models can be found in the Supplementary Material (Figures~\ref{figure:10_8_models_Optuna_and_two_FT_lines_1} - \ref{figure:10_8_models_Optuna_and_two_FT_lines_3}).
\section{Conclusion}

Our findings demonstrate that the fine-tuned Code Llama is highly effective in identifying optimal hyperparameter values across diverse machine learning models, achieving improvements in both accuracy and stability. The fine-tuned model not only enhances training efficiency, but also delivers competitive performance in various scenarios, making it a valuable asset for researchers and practitioners.

Our evaluations reveal that fine-tuned Code Llama often meets or exceeds the accuracy achieved by Optuna, a well-established hyperparameter optimization framework. In particular, fine-tuning cycles resulted in higher accuracy and more stable predictions across multiple models, including complex architectures such as ResNet and DenseNet, where fine-tuning significantly surpassed baseline results. These architectures are commonly used in image restoration and manipulation tasks, highlighting the practical relevance of our method to real-world vision applications.

Additionally, one-shot predictions using Code Llama demonstrated accuracy levels close to those obtained through full fine-tuning, highlighting its potential for rapid and effective optimization.

A key insight from this study is the efficiency advantage of fine-tuned Code Llama. Although Optuna typically requires training across 100 trials to identify suitable hyperparameters, our approach achieves comparable or superior results with significantly fewer computational resources. This makes the method especially attractive for image enhancement workflows, where perceptual quality and resource efficiency must be balanced.

In general, our study underscores the effectiveness of fine-tuned Code Llama as a powerful tool for hyperparameter optimization. Its ability to provide stable high-quality results with minimal training iterations positions it as a compelling alternative to traditional optimization methods, contributing to the development of smarter, faster, and more adaptive pipelines for image manipulation and restoration.
\vspace{0.2cm}

\noindent\textbf{Acknowledgments.}
This work was partially supported by the Alexander von Humboldt Foundation. 
{
    \small
    \bibliographystyle{ieeenat_fullname}
    \bibliography{main}
}
\clearpage
\setcounter{page}{1}
\maketitlesupplementary

\section{RMSE Comparison}

\subsection{Models from train dataset}

\begin{table*}[ht]
\centering
\begin{tabular}{lcccccc}
\toprule
\textbf{Model} & \textbf{Epochs} & \textbf{RMSE Optuna} & \makecell{\textbf{RMSE}\\\textbf{FT Cycle 1}} & \makecell{\textbf{Difference }\\\textbf{Cycle 1}} & \makecell{\textbf{RMSE}\\\textbf{FT Cycle 2}} & \makecell{\textbf{Difference }\\\textbf{Cycle 2}} \\
\hline
AlexNet         & 1 & 0.8258 & 0.8421 & -0.0163 & 0.5779 & 0.2479 \\
AlexNet         & 2 & 0.8236 & 0.6912 & 0.1324 & 0.4221 & 0.4015 \\
ConvNeXt        & 1 & 0.7924 & 0.7460 & 0.0464 & 0.7493 & 0.0431 \\
ConvNeXt        & 2 & 0.7426 & 0.6861 & 0.0565 & 0.6863 & 0.0563 \\
DenseNet        & 1 & 0.4930 & 0.4526 & 0.0404 & 0.5899 & -0.0969 \\
DenseNet        & 2 & 0.3525 & 0.3325 & 0.0201 & 0.4307 & -0.0782 \\
EfficientNet    & 1 & 0.4835 & 0.6250 & -0.1415 & 0.6498 & -0.1663 \\
EfficientNet    & 2 & 0.3370 & 0.5486 & -0.2116 & 0.5563 & -0.2193 \\
GoogLeNet       & 1 & 0.5158 & 0.5865 & -0.0707 & 0.5093 & 0.0065 \\
GoogLeNet       & 2 & 0.3956 & 0.3857 & 0.0099 & 0.3044 & 0.0912 \\
MNASNet         & 1 & 0.4613 & 0.5309 & -0.0696 & 0.5132 & -0.0519 \\
MNASNet         & 2 & 0.3682 & 0.4308 & -0.0626 & 0.4620 & -0.0938 \\
MobileNetV2     & 1 & 0.4654 & 0.4155 & 0.0499 & 0.4316 & 0.0338 \\
MobileNetV2     & 2 & 0.3617 & 0.2854 & 0.0763 & 0.4279 & -0.0662 \\
MobileNetV3     & 1 & 0.4852 & 0.7328 & -0.2476 & 0.8186 & -0.3334 \\
MobileNetV3     & 2 & 0.3368 & 0.6797 & -0.3429 & 0.6567 & -0.3199 \\
RegNet          & 1 & 0.5677 & 0.5392 & 0.0285 & 0.6254 & -0.0577 \\
RegNet          & 2 & 0.4368 & 0.3865 & 0.0503 & 0.5484 & -0.1116 \\
ResNet          & 1 & 0.4692 & 0.5153 & -0.0461 & 0.5847 & -0.1155 \\
ResNet          & 2 & 0.3371 & 0.4325 & -0.0954 & 0.5601 & -0.2230 \\
ShuffleNet      & 1 & 0.4832 & 0.4282 & 0.0550 & 0.5222 & -0.0390 \\
ShuffleNet      & 2 & 0.3471 & 0.3236 & 0.0235 & 0.4638 & -0.1167 \\
SqueezeNet      & 1 & 0.8597 & 0.7200 & 0.1397 & 0.6888 & 0.1709 \\
SqueezeNet      & 2 & 0.8532 & 0.6676 & 0.1856 & 0.5829 & 0.2703 \\
SwinTransformer & 1 & 0.7500 & 0.6527 & 0.0973 & 0.6746 & 0.0754 \\
SwinTransformer & 2 & 0.6726 & 0.5676 & 0.1050 & 0.6237 & 0.0489 \\
VGG             & 1 & 0.8192 & 0.5898 & 0.2295 & 0.5364 & 0.2829 \\
VGG             & 2 & 0.7961 & 0.3741 & 0.4220 & 0.3743 & 0.4218 \\
\bottomrule
\end{tabular}
\caption{Comparison of RMSE differences for fine-tuning cycles 1 and 2 for each computer vision model and epoch from the training dataset. The table highlights the RMSE values obtained using Optuna as a baseline (RMSE Optuna) and compares them with the RMSE values achieved during the first and second fine-tuning cycles of Code Llama (RMSE FT Cycle 1 and RMSE FT 2, respectively). The Difference Cycle 1 and Difference Cycle 2 columns indicate the changes in RMSE relative to Optuna for the first and second fine-tuning cycles, respectively. Positive values in the difference columns represent improvements, while negative values indicate an increase in RMSE.}
\label{table:rmse_comparison_train_cv_combined}
\end{table*}

\begin{table}[h]
\centering
\begin{tabular}{lcccc}
\toprule
% \multicolumn{2}{|c|}{\textbf{Fine-tuning cycle 1}} \\ \hline
\textbf{Model} & \textbf{Epochs} & \makecell{\textbf{RMSE}\\\textbf{Optuna}} & \makecell{\textbf{RMSE}\\\textbf{FT 1}} & \textbf{Difference} \\
\hline 
RNN     & 1 & 0.7959 & 0.7880 & 0.0079 \\
RNN     & 2 & 0.7888 &  0.7703 & 0.0185 \\
LSTM    & 1 & 0.7691 & 0.7564 & 0.0127 \\
LSTM    & 2 & 0.7468 & 0.7478 & -0.0010 \\
Llama   & 1 & 0.9988 & 0.9990 & -0.0002 \\
Llama   & 2 & 0.9987 & 0.9991 & -0.0004 \\
\bottomrule
\end{tabular}
\caption{Comparison of RMSE values for text generation models from the training dataset during fine-tuning cycle 1. The table highlights the differences in RMSE between the baseline results (Optuna) and the fine-tuned results (FT 1). Positive differences indicate an improvement in RMSE after fine-tuning, while negative differences suggest a potential decline in performance.}
\label{table:rmse_comparison_train_text_generation}
\end{table}

\begin{table*}[h]
\centering
\begin{tabular}{lcccc}
\toprule
\textbf{Model} & \textbf{Epochs} & \textbf{RMSE Optuna} & \textbf{RMSE FT 2} & \textbf{Difference} \\ 
\hline
InceptionV3       & 1 & 0.5906 & 0.8269 & -0.2363 \\
InceptionV3       & 2 & 0.4323 & 0.7687 & -0.3364 \\
MaxVit            & 1 & 0.8349 & 0.6430 & 0.1919 \\
MaxVit            & 2 & 0.8297 & 0.5356 & 0.2941 \\
VisionTransformer & 1 & 0.7768 & 0.5884 & 0.1884 \\
VisionTransformer & 2 & 0.7659 & 0.5253 & 0.2406 \\
\bottomrule
\end{tabular}
\caption{Comparison of RMSE differences between Optuna and Fine-tuning Cycle 2 for models from the test dataset. Positive differences indicate an improvement in RMSE after fine-tuning, while negative differences suggest a potential decline in performance.}
\label{table:rmse_comparison_test_cv}
\end{table*}

{\textbf{Computer vision models.}}
Table \ref{table:rmse_comparison_train_cv_combined} presents a detailed comparison of RMSE values for each model and epoch under three conditions: baseline Optuna, Code Llama after the first fine-tuning cycle (FT Cycle 1), and Code Llama after the second fine-tuning cycle (FT Cycle 2). The final two columns (\emph{Difference 1} and \emph{Difference 2}) show how much RMSE changes relative to the baseline, with positive values indicating that Code Llama yields a \emph{lower} RMSE than Optuna (i.e., an improvement), and negative values implying the baseline outperforms Code Llama.

A number of architectures exhibit immediate gains in the first fine-tuning cycle. For example, DenseNet, MobileNetV2, RegNet, ShuffleNet, SqueezeNet, SwinTransformer, and VGG consistently show positive differences across their epochs, indicating a clear reduction in RMSE relative to Optuna. ConvNeXt also demonstrates improvements in both epochs, confirming Code Llama’s ability to quickly adapt to a variety of high-performance models. AlexNet shows a notable positive difference in epoch 2, reflecting early and meaningful gains in hyperparameter recommendations. In contrast, some models — such as EfficientNet and GoogLeNet (epoch 1) — register negative differences, implying that for certain configurations, Optuna remains more effective in finding lower-RMSE settings.

The second fine-tuning cycle further improves RMSE in several models. AlexNet continues to perform well, with both epochs showing additional gains over the first cycle. Similarly, GoogLeNet and SqueezeNet exhibit further improvements across both epochs, highlighting the benefits of iterative tuning. In models such as ConvNeXt and MobileNetV3 (epoch 2), the second cycle also yields enhanced performance compared to the first, mitigating earlier shortcomings.

However, not all models benefit equally from iterative LLM-based fine-tuning. For instance, MNASNet shows a drop in performance in the second epoch, indicating a potential limitation in its adaptability to the fine-tuning process. Similarly, ResNet and RegNet show a decline in RMSE improvements in the second cycle, pointing to possible instability in the learned hyperparameter patterns. These observations highlight the importance of carefully assessing trade-offs and potential limitations when applying fine-tuning to specific model types.

When comparing the fine-tuning results to the Optuna baseline, it becomes clear that fine-tuning effectively reduces RMSE for the majority of models and epochs — particularly for computationally demanding architectures, such as ConvNeXt, SwinTransformer, and VGG. These results affirm that fine-tuning is a robust optimization strategy capable of surpassing or complementing traditional hyperparameter search methods like Optuna for many computer vision tasks. At the same time, the variability across models and epochs underscores the need for architecture-specific tuning strategies to maximize the benefits of fine-tuning.

\paragraph{\textbf{Text generation models.}}
The analysis of RMSE differences for the first fine-tuning cycle, as shown in Table~\ref{table:rmse_comparison_train_text_generation}, provides detailed insight into the impact of fine-tuning on text generation models. For the RNN model, fine-tuning results in a small but consistent improvement in RMSE—approximately 0.0185 lower than the Optuna baseline at epoch 2 — indicating stable and beneficial adjustments from the LLM.

The LSTM model also shows improvement, reducing RMSE by about 0.0127 at epoch 1 compared to Optuna, although a minor performance drop is observed at epoch 2. This variation suggests that LSTM responds positively to fine-tuning overall, though tuning depth may affect consistency.

In contrast, the Llama model shows a slight decline in performance across both epochs, with RMSE values marginally worse than Optuna. This may point to the challenges of adapting large, complex architectures like Llama using the current tuning strategy.

\subsection{Models from test dataset}

The analysis of the RMSE differences between Optuna and the second fine-tuning cycle for models from the test dataset, as shown in Table \ref{table:rmse_comparison_test_cv}, evaluates the impact of fine-tuning on previously unseen computer vision models.

Among the evaluated models, MaxVit and VisionTransformer show substantial improvements after fine-tuning. MaxVit achieves positive RMSE differences of 0.1919 and 0.2941 for epochs 1 and 2, respectively, while VisionTransformer improves by 0.1884 and 0.2406. These results indicate that the fine-tuning process significantly enhances the accuracy of these architectures, even though they were not part of the original training set. Such outcomes suggest that Code Llama generalizes well to new models, successfully adapting its hyperparameter recommendations to unfamiliar architectures.

In contrast, InceptionV3 demonstrates a notable decline in performance following the second fine-tuning cycle. RMSE values increase by –0.2363 and -0.3364 for epochs 1 and 2, respectively, compared to the Optuna baseline.   While these results indicate that the current fine-tuning setup may not fully optimize InceptionV3, they also point to valuable opportunities for future enhancement. With refined prompt design, adjusted parameter initialization, or extended fine-tuning schedules, there is strong potential to better align the model’s architecture with the optimization process and achieve improved outcomes.

\section{Best Accuracy Comparison}

\subsection{Models from train dataset}

Table \ref{table:best_accuracies_train_combined} presents a detailed result of the best accuracy achieved by Optuna and fine-tuned Code Llama across two fine-tuning cycles for a diverse range of computer vision models and epochs.

The fine-tuning process demonstrates its ability to enhance accuracy for many models, with notable improvements observed during the second cycle. For instance, MNASNet and MobileNetV2 achieve significant gains in the first epoch of the second fine-tuning cycle, underscoring the effectiveness of fine-tuning in refining hyperparameters to better align with the models’ architectures. Similarly, RegNet and SqueezeNet show consistent accuracy improvements across cycles, particularly excelling in the second cycle, showcasing the iterative benefits of Code Llama’s fine-tuning framework.

The results also highlight the stability achieved through fine-tuning, as models such as DenseNet and ResNet demonstrate consistent accuracy across multiple epochs and cycles. This stability reflects the robustness of Code Llama’s fine-tuning methodology, positioning it as a reliable tool for hyperparameter optimization. In many cases, the fine-tuning process achieves accuracy results comparable to or exceeding those of Optuna, reinforcing its utility as a competitive alternative.

While certain models, such as MobileNetV3 and SwinTransformer, exhibit moderate or variable improvements across cycles, these findings underscore the importance of tailoring fine-tuning strategies to the specific requirements of each architecture. For models that are well-optimized in the initial cycle, additional fine-tuning may yield diminishing returns, suggesting the need for a balanced approach that considers computational costs alongside expected performance gains. This variability highlights the importance of model-specific strategies in leveraging the full potential of fine-tuning.

\begin{table*}[h]
    \centering
    \begin{tabular}{lcccccc}
        \toprule
        % \textbf{Model} & \textbf{Epochs} & \textbf{Best acc Optuna} & \textbf{Best acc FT 1} & \textbf{Difference 1} & \textbf{Best acc FT 2} & \textbf{Difference 2} \\
        \textbf{Model} & \textbf{Epochs} & \makecell{\textbf{Best Accuracy}\\\textbf{Optuna}} & \makecell{\textbf{Best Accuracy}\\\textbf{FT Cycle 1}} & \makecell{\textbf{Difference}\\\textbf{Cycle 1}} & \makecell{\textbf{Best Accuracy}\\\textbf{FT Cycle 2}} & \makecell{\textbf{Difference}\\\textbf{Cycle 2}} \\
        \hline
        MNASNet        & 1 & 0.6449 & 0.6390 & -0.0059 & 0.7184 & 0.0735 \\
        MNASNet        & 2 & 0.7480 & 0.7476 & -0.0004 & 0.7420 & -0.0060 \\
        MobileNetV2    & 1 & 0.6452 & 0.6424 & -0.0028 & 0.7365 & 0.0913 \\
        MobileNetV2    & 2 & 0.7493 & 0.7457 & -0.0036 & 0.7446 & -0.0047 \\
        AlexNet        & 1 & 0.5364 & 0.5392 & 0.0028  & 0.6282 & 0.0918 \\
        AlexNet        & 2 & 0.6640 & 0.6623 & -0.0017 & 0.6680 & 0.0040 \\
        MobileNetV3    & 1 & 0.6322 & 0.6228 & -0.0094 & 0.6174 & -0.0148 \\
        MobileNetV3    & 2 & 0.7488 & 0.7314 & -0.0174 & 0.7149 & -0.0339 \\
        ConvNeXt       & 1 & 0.3454 & 0.3393 & -0.0061 & 0.3020 & -0.0434 \\
        ConvNeXt       & 2 & 0.4085 & 0.3962 & -0.0123 & 0.4119 & 0.0034 \\
        VGG            & 1 & 0.5829 & 0.5859 & 0.0030  & 0.5804 & -0.0025 \\
        VGG            & 2 & 0.7031 & 0.6828 & -0.0203 & 0.6973 & -0.0058 \\
        SwinTransformer & 1 & 0.4548 & 0.4365 & -0.0183 & 0.4345 & -0.0203 \\
        SwinTransformer & 2 & 0.5311 & 0.5180 & -0.0131 & 0.5029 & -0.0282 \\
        DenseNet       & 1 & 0.6316 & 0.6177 & -0.0139 & 0.6216 & -0.0100 \\
        DenseNet       & 2 & 0.7438 & 0.7420 & -0.0018 & 0.7461 & 0.0023 \\
        SqueezeNet     & 1 & 0.4017 & 0.3848 & -0.0169 & 0.4006 & -0.0011 \\
        SqueezeNet     & 2 & 0.4742 & 0.4662 & -0.0080 & 0.4820 & 0.0078 \\
        EfficientNet   & 1 & 0.6170 & 0.6198 & 0.0028  & 0.5935 & -0.0235 \\
        EfficientNet   & 2 & 0.7374 & 0.7387 & 0.0013  & 0.7037 & -0.0337 \\
        GoogLeNet      & 1 & 0.6405 & 0.6746 & 0.0341  & 0.6538 & 0.0133 \\
        GoogLeNet      & 2 & 0.7494 & 0.7402 & -0.0092 & 0.7630 & 0.0136 \\
        ShuffleNet     & 1 & 0.6346 & 0.6369 & 0.0023  & 0.6321 & -0.0025 \\
        ShuffleNet     & 2 & 0.7185 & 0.7246 & 0.0061  & 0.7177 & -0.0008 \\
        RegNet         & 1 & 0.5289 & 0.5331 & 0.0042  & 0.5466 & 0.0177 \\
        RegNet         & 2 & 0.6498 & 0.6523 & 0.0025  & 0.6629 & 0.0131 \\
        ResNet         & 1 & 0.6255 & 0.6241 & -0.0014 & 0.6282 & 0.0027 \\
        ResNet         & 2 & 0.7378 & 0.7399 & 0.0021  & 0.7399 & 0.0021 \\
        \bottomrule
    \end{tabular}
    \caption{Comparison of the best accuracy differences for fine-tuning cycles 1 and 2 across various computer vision models and epochs, alongside the best accuracy achieved using Optuna. Each row represents a model evaluated for 1 or 2 epochs, comparing the performance of fine-tuned models (FT Cycle 1 and FT Cycle 2) with the baseline Optuna results. Positive values in the Difference columns (Difference Cycle 1 and Difference Cycle 2) indicate an improvement in accuracy after fine-tuning, while negative values suggest a decrease in performance.}
    \label{table:best_accuracies_train_combined}
    \vspace{+5mm}
\end{table*}

\subsection{Models from test dataset}
The results presented in Table \ref{table:best_accuracies_test} provide a detailed evaluation of the accuracy differences between the baseline Optuna approach and the second fine-tuning cycle (FT Cycle 2) for models in the test dataset.

The fine-tuning process introduces significant changes in model accuracy, with varying results depending on the architecture. For InceptionV3, fine-tuning in FT Cycle 2 results in a notable reduction in accuracy compared to Optuna, with differences of -0.3001 for one epoch and -0.2423 for two epochs. These findings suggest that while fine-tuning significantly adjusts the model’s parameters, it may not always align with the specific optimization needs of complex architectures like InceptionV3. This highlights the potential necessity of more customized hyperparameter tuning strategies for such models to better leverage fine-tuning.

For VisionTransformer, the differences between Optuna and FT Cycle 2 are less pronounced, with values of -0.0523 for one epoch and -0.0427 for two epochs. This stability in performance suggests that the second cycle of fine-tuning effectively preserves the model’s baseline accuracy while introducing modest refinements. The relatively small variations indicate that VisionTransformer may require fewer fine-tuning adjustments to reach or maintain optimal performance, reflecting its robustness to the fine-tuning process.

MaxVit demonstrates the most consistent and positive results among the evaluated models. The difference is minimal for one epoch (-0.0018), and for two epochs, FT Cycle 2 outperforms Optuna with a positive difference of 0.0249. This improvement underscores MaxVit’s adaptability to fine-tuning, highlighting its potential for further optimization.

\begin{table*}[h]
\centering
\begin{tabular}{lcccc}
\toprule
\textbf{Model} & \textbf{Epochs} & \makecell{\textbf{Best Accuracy}\\\textbf{Optuna}} & \makecell{\textbf{Best Accuracy}\\\textbf{FT Cycle 2}} & \textbf{Difference} \\
\hline
InceptionV3 & 1 & 0.5137 & 0.2136 & -0.3001 \\
InceptionV3 & 2 & 0.6818 & 0.4395 & -0.2423 \\
VisionTransformer & 1 & 0.4715 & 0.4192 & -0.0523 \\
VisionTransformer & 2 & 0.533 & 0.4903 & -0.0427 \\
MaxVit & 1 & 0.4323 & 0.4305 & -0.0018 \\
MaxVit & 2 & 0.5096 & 0.5345 & 0.0249 \\
\bottomrule
\end{tabular}
\caption{Comparison of the best accuracy between Optuna and Fine-tuned (FT Cycle 2) models for test dataset architectures. Each row represents a specific model evaluated after 1 or 2 epochs, comparing the best accuracy achieved using Optuna with the accuracy obtained after the second cycle of fine-tuning (FT Cycle 2). The Difference column highlights the variation in performance, where positive values indicate an improvement in accuracy after fine-tuning, and negative values suggest a decline.}
\label{table:best_accuracies_test}
\end{table*}

\section{Prediction Dynamics Across Epochs}
\begin{figure*}[p]
    \centering
    \setlength{\tabcolsep}{4pt}
    
    \renewcommand{\arraystretch}{10.0}
    \begin{tabular}{cc}
        
        % Line 1
        \begin{minipage}{0.45\textwidth}
            \centering
            \includegraphics[width=\linewidth]{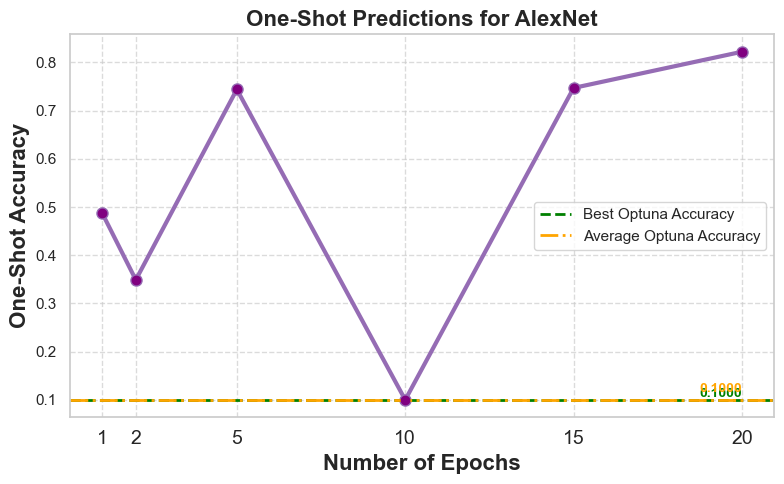}
        \end{minipage} &
        \begin{minipage}{0.45\textwidth}
            \centering
            \includegraphics[width=\linewidth]{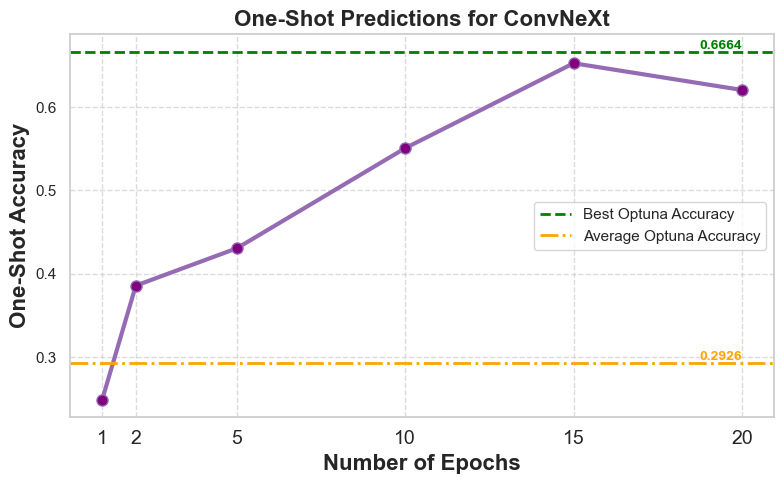}
        \end{minipage} \\
        
        % Line 2
        \begin{minipage}{0.45\textwidth}
            \centering
            \includegraphics[width=\linewidth]{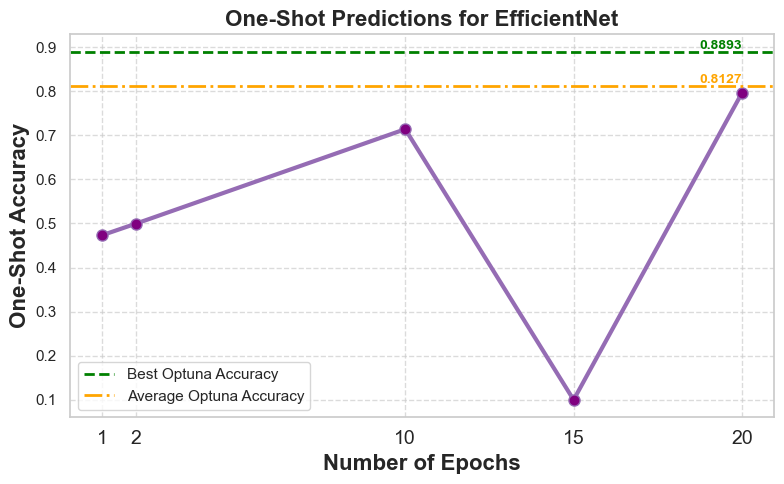}
            % \captionof{figure}{Image 3}
        \end{minipage} &
        \begin{minipage}{0.45\textwidth}
            \centering
            \includegraphics[width=\linewidth]{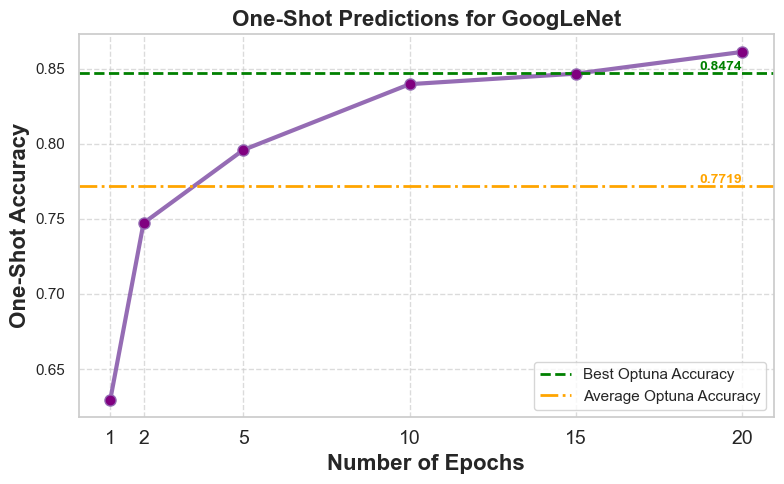}
            % \captionof{figure}{Image 4}
        \end{minipage} \\

        % Line 3
        \begin{minipage}{0.45\textwidth}
            \centering
            \includegraphics[width=\linewidth]{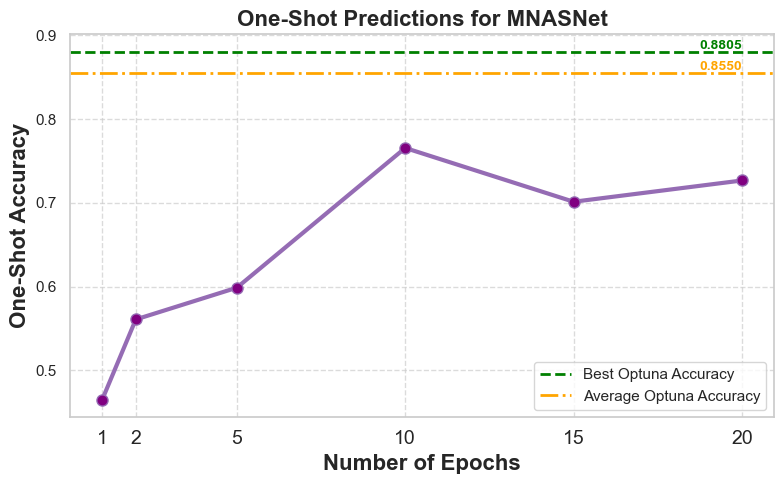}
            % \captionof{figure}{Image 3}
        \end{minipage} &
        \begin{minipage}{0.45\textwidth}
            \centering
            \includegraphics[width=\linewidth]{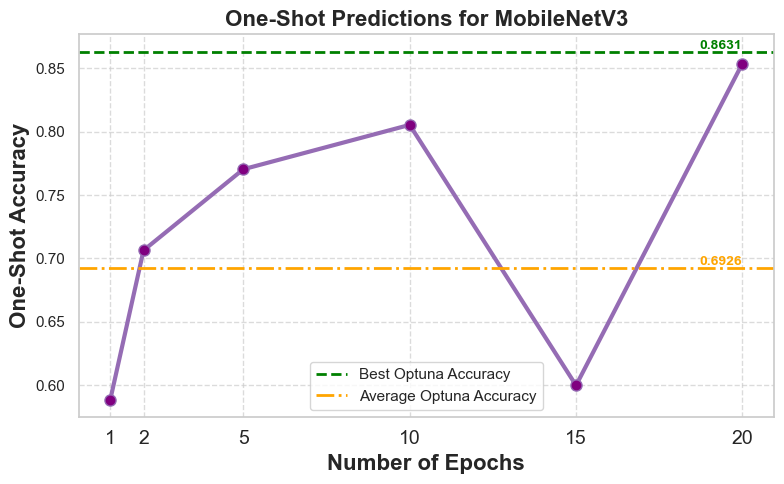}
            % \captionof{figure}{Image 4}
        \end{minipage} \\

        % Line 4
        \begin{minipage}{0.45\textwidth}
            \centering
            \includegraphics[width=\linewidth]{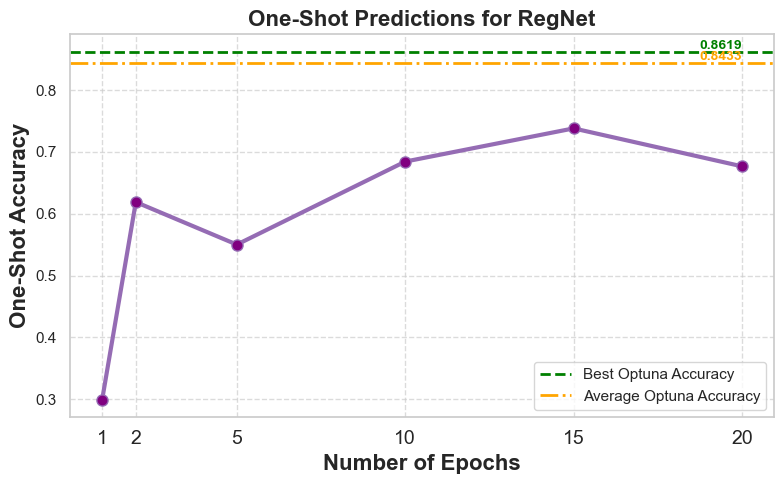}
            % \captionof{figure}{Image 3}
        \end{minipage} &
        \begin{minipage}{0.45\textwidth}
            \centering
            \includegraphics[width=\linewidth]{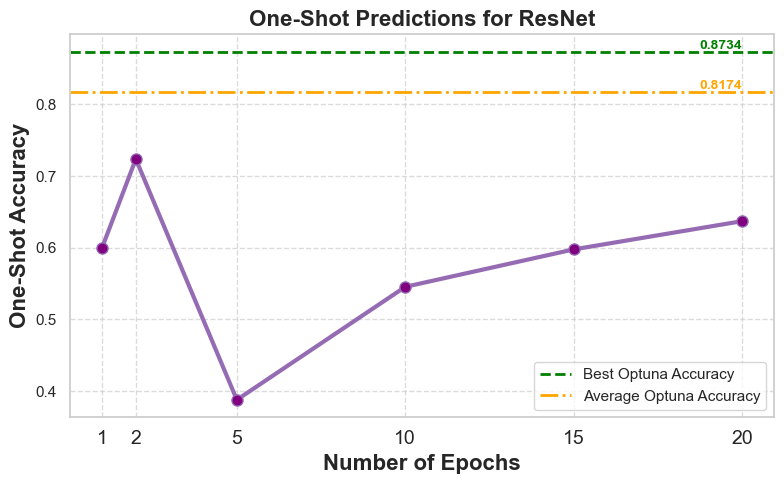}
            % \captionof{figure}{Image 4}
        \end{minipage} \\
    \end{tabular}
\caption{Part 1: Accuracy trends observed in one-shot predictions for 13 computer vision models over varying epochs (1, 2, 5, 10, 15, and 20). Each plot represents the accuracy performance of a specific model using the one-shot prediction approach. All models in this part are derived from the train dataset, demonstrating the trends within the training data context.} 
\label{figure:11_13_models_One_Shot_1}
\end{figure*}

\begin{figure*}[p]
    \centering
    \setlength{\tabcolsep}{4pt}
    
    \renewcommand{\arraystretch}{10.0}
    \begin{tabular}{cc}

        % Line 5
        \begin{minipage}{0.45\textwidth}
            \centering
            \includegraphics[width=\linewidth]{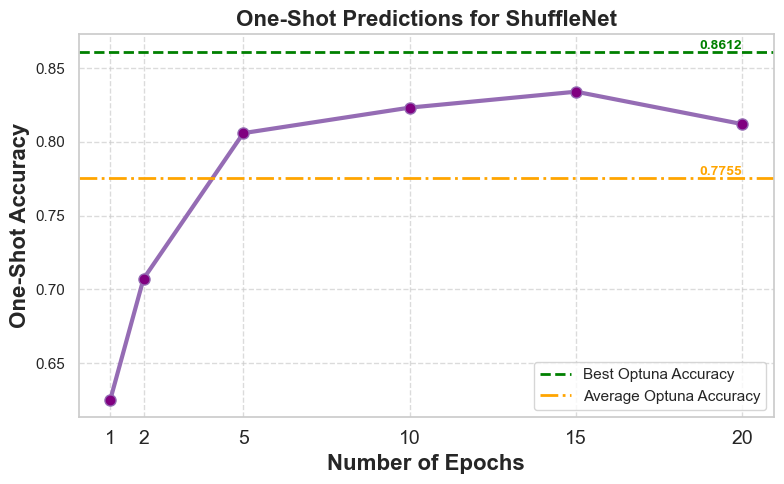}
            % \captionof{figure}{Image 3}
        \end{minipage} &
        \begin{minipage}{0.45\textwidth}
            \centering
            \includegraphics[width=\linewidth]{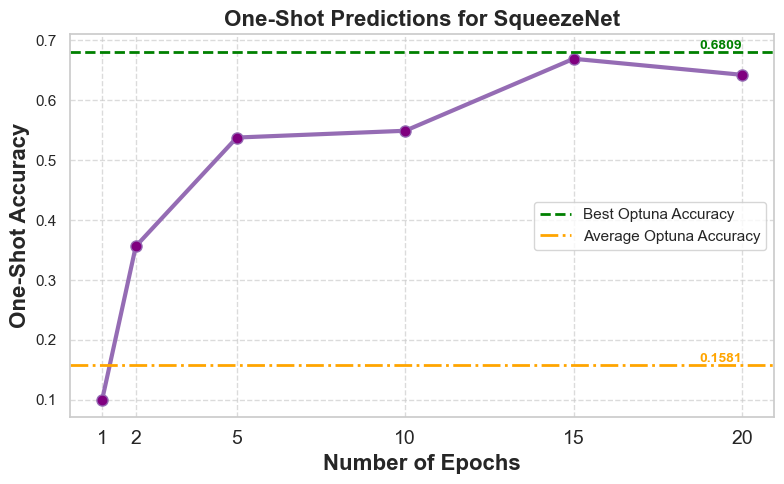}
            % \captionof{figure}{Image 4}
        \end{minipage} \\

        % Line 6
        \begin{minipage}{0.45\textwidth}
            \centering
            \includegraphics[width=\linewidth]{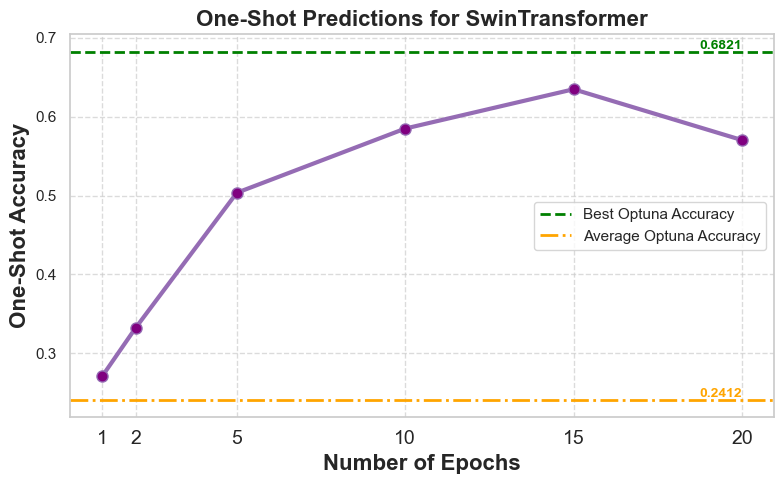}
            % \captionof{figure}{Image 3}
        \end{minipage} &
        \begin{minipage}{0.45\textwidth}
            \centering
            \includegraphics[width=\linewidth]{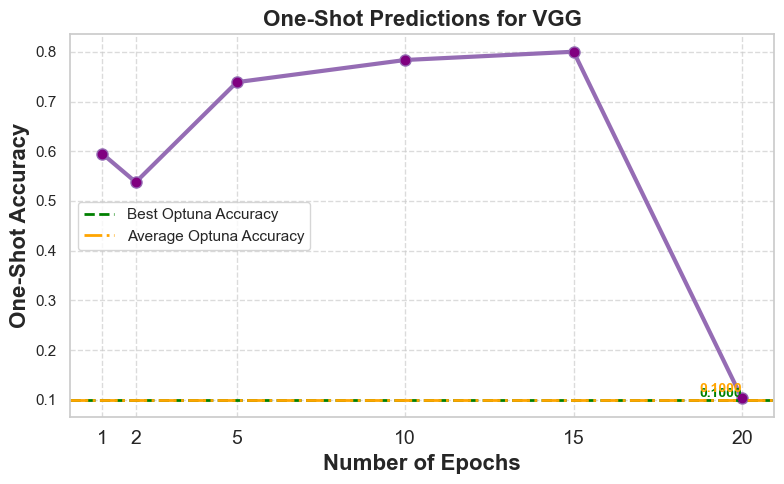}
            % \captionof{figure}{Image 4}
        \end{minipage} \\

                % Line 7
        \multicolumn{2}{c}{
            \begin{minipage}{0.45\textwidth}
                \centering
                \includegraphics[width=\linewidth]{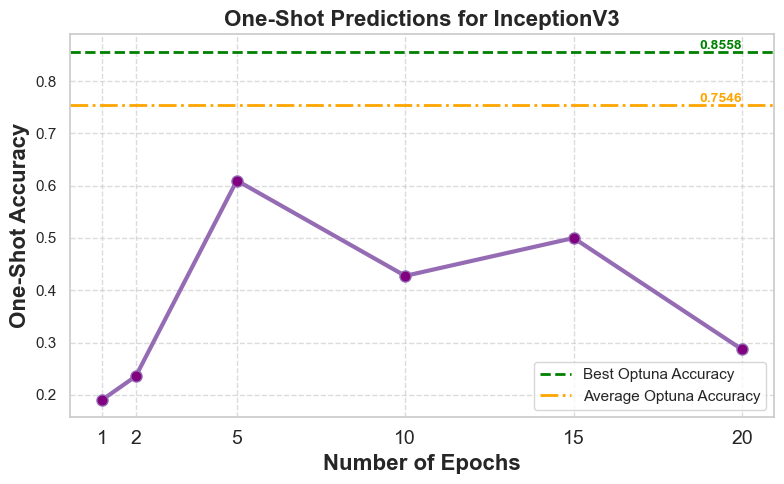}
                % \captionof{figure}{Image 3}
            \end{minipage}
        } \\
    \end{tabular}
\caption{Part 2: Accuracy trends observed in one-shot predictions for 13 computer vision models over varying epochs (1, 2, 5, 10, 15, and 20). Each plot represents the accuracy performance of a specific model using the one-shot prediction approach. All models in this part are derived from the train dataset, except for InceptionV3, which is taken from the test dataset and was not involved in the fine-tuning of Code Llama.} 
\label{figure:11_13_models_One_Shot_2}
\end{figure*}

The results of one-shot predictions across 13 computer vision models, which are not covered in Section \ref{subsection:Prediction_Dynamics_Across_Epochs} of this paper, for different numbers of epochs (1, 2, 5, 10, 15, and 20), shown in Figures \ref{figure:11_13_models_One_Shot_1}-\ref{figure:11_13_models_One_Shot_2}, demonstrate a clear trend of increasing accuracy with additional training epochs for these models. This improvement reflects the expected behavior, as extended training allows models to refine their predictions and optimize performance. For many architectures, including ConvNeXt, ShuffleNet, and ResNet, accuracy stabilizes after 10 epochs, indicating that these models reach their optimal performance within this training range.

Interestingly, models such as GoogLeNet, ShuffleNet, and DenseNet exhibit high accuracy even at early epochs, showcasing their ability to converge quickly and effectively. These results highlight their potential for efficient training in scenarios where computational resources or time are limited. Additionally, one-shot predictions for several models, such as MobileNetV3 and EfficientNet, reveal that their learning dynamics can vary across epochs, reflecting the unique characteristics of their architectures and the data.

However, not all models exhibit consistent improvements in accuracy. For instance, MobileNetV3 shows a significant drop in performance after 15 epochs before recovering at 20 epochs, suggesting potential overfitting or instability during certain training phases. Similarly, InceptionV3, which originates from the test dataset and was not included in the fine-tuning process of Code Llama, struggles to maintain accuracy gains as training progresses, with a notable decline in performance by the 20th epoch. These results indicate that certain architectures, particularly those not exposed to fine-tuning, may face challenges in sustaining accuracy improvements across extended training, potentially requiring more tailored hyperparameter adjustments or regularization strategies.

Across all models, it is evident that fine-tuned hyperparameters provided by Code Llama contribute to strong prediction performance. Notably, GoogLeNet, ShuffleNet, and ResNet display consistently high stability and accuracy across the evaluated epochs, emphasizing their robustness in various training settings. Even for architectures that show more variability, such as MobileNetV3 and MNASNet, the overall trend demonstrates the ability of Code Llama to enhance model performance efficiently.

These findings demonstrate the value of one-shot predictions and emphasize the need to account for the specific characteristics of each model when designing training strategies. By using fine-tuned hyperparameters and adjusting training durations to suit the unique requirements of each architecture, it becomes possible to achieve high accuracy while efficiently utilizing computational resources.

\section{Training Trends and Performance Comparison}

\begin{figure*}[h!]
    \centering
    \setlength{\tabcolsep}{4pt}
    \renewcommand{\arraystretch}{10.0}
    \begin{tabular}{cc}
        
        % Line 1
        \begin{minipage}{0.45\textwidth}
            \centering
            \includegraphics[width=\linewidth]{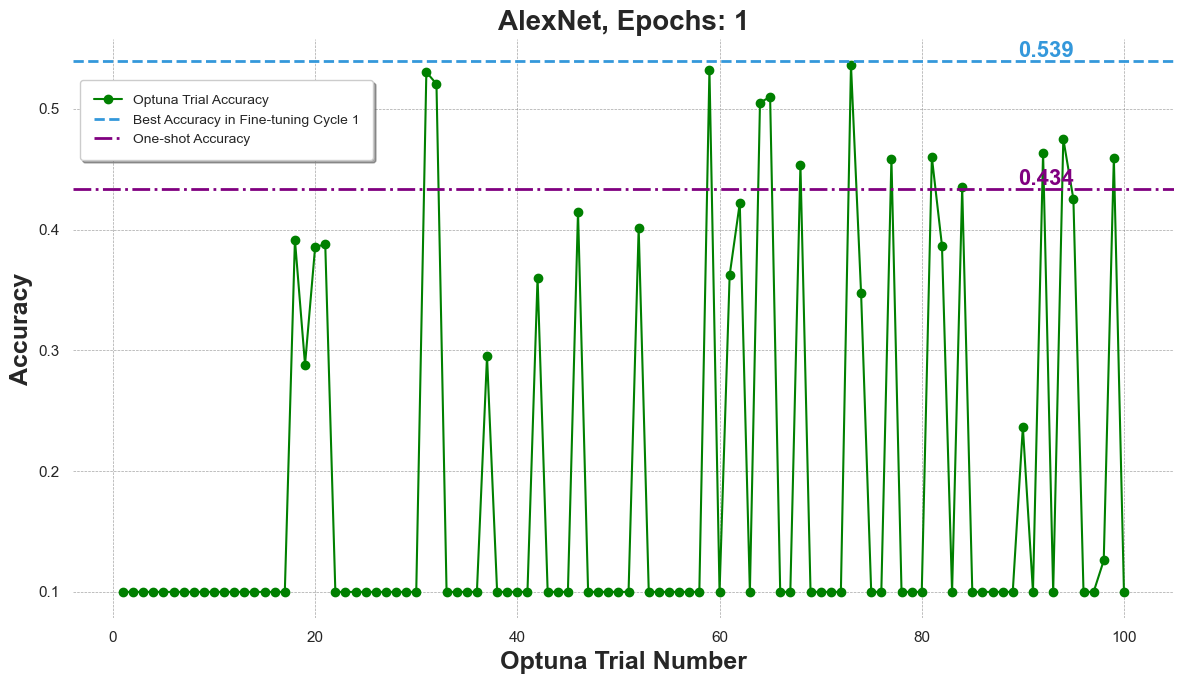}
            % \captionof{figure}{Image 1}
        \end{minipage} &
        \begin{minipage}{0.45\textwidth}
            \centering
            \includegraphics[width=\linewidth]{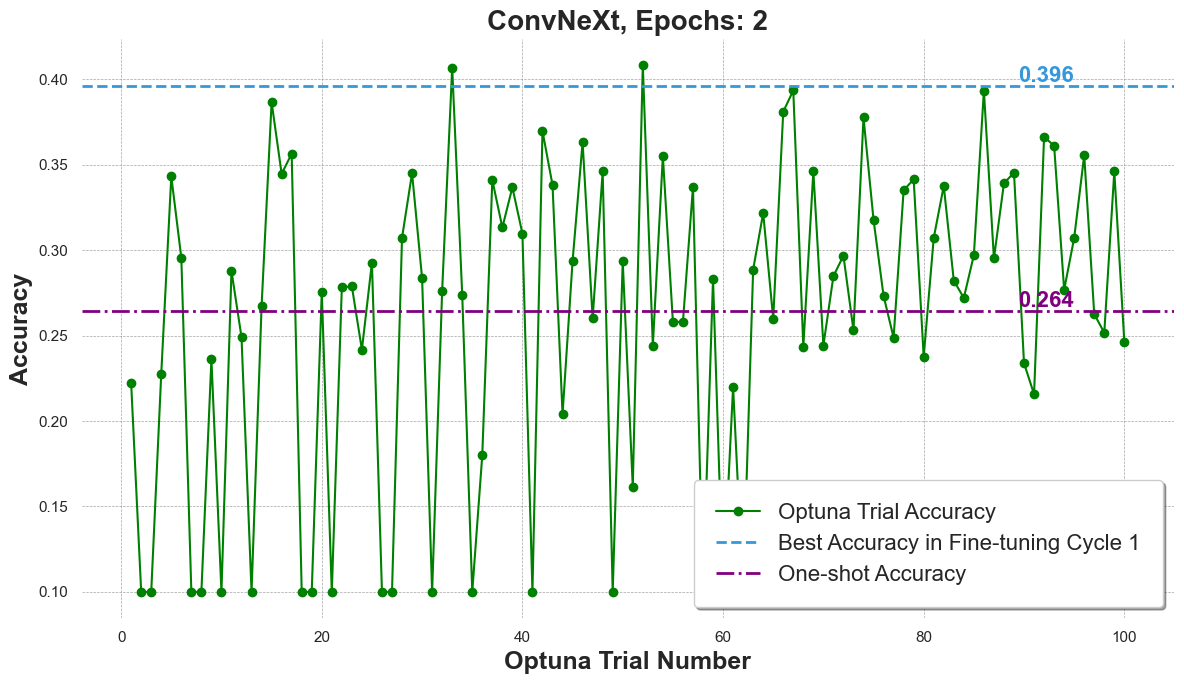}
            % \captionof{figure}{Image 2}
        \end{minipage} \\
        
        % Line 3
        \begin{minipage}{0.45\textwidth}
            \centering
            \includegraphics[width=\linewidth]{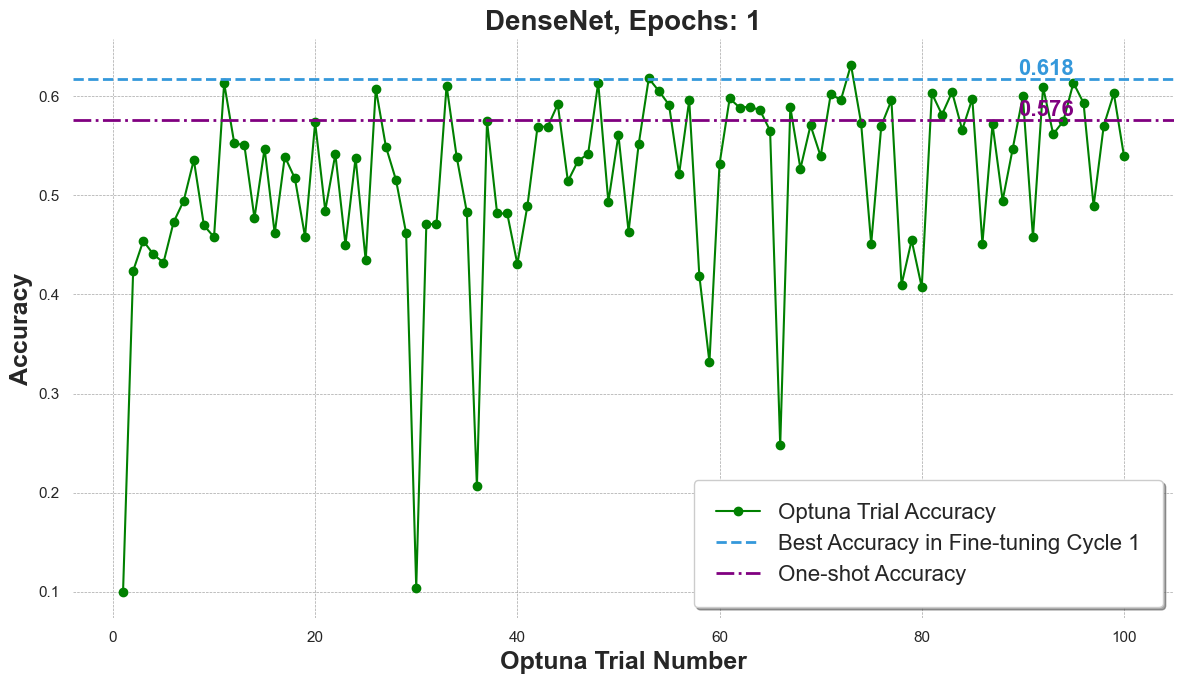}
            % \captionof{figure}{Image 3}
        \end{minipage} &
        \begin{minipage}{0.45\textwidth}
            \centering
            \includegraphics[width=\linewidth]{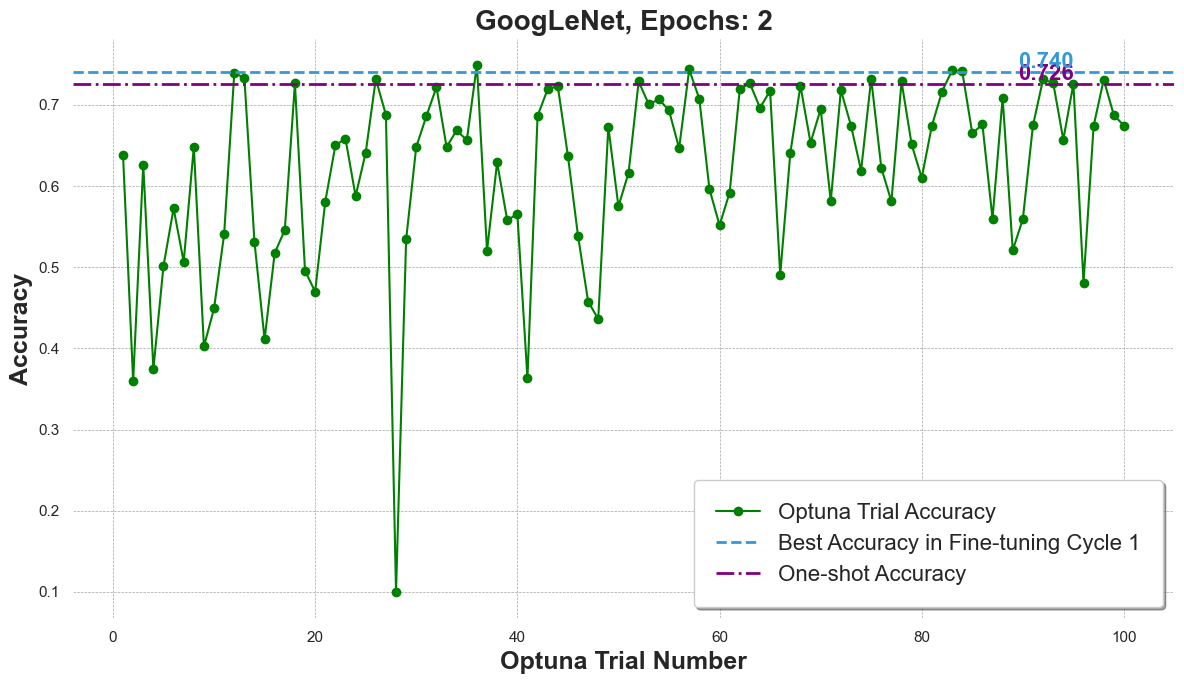}
            % \captionof{figure}{Image 4}
        \end{minipage} \\

        % Line 4
        \begin{minipage}{0.45\textwidth}
            \centering
            \includegraphics[width=\linewidth]{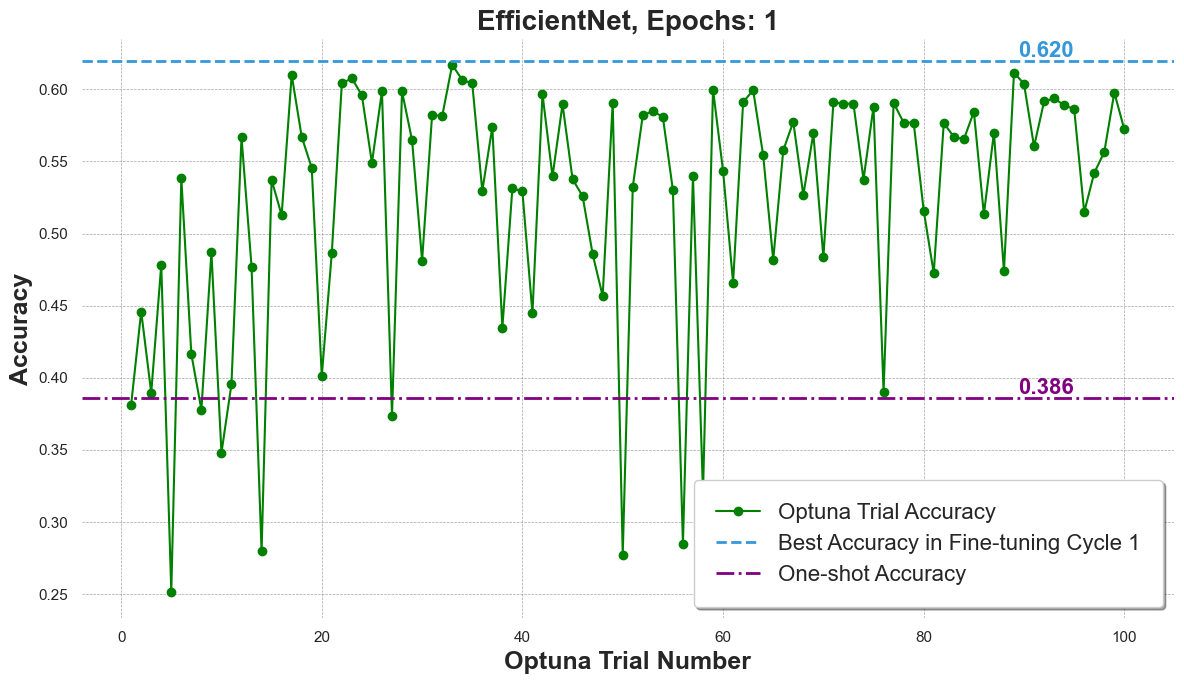}
            % \captionof{figure}{Image 3}
        \end{minipage} &
        \begin{minipage}{0.45\textwidth}
            \centering
            \includegraphics[width=\linewidth]{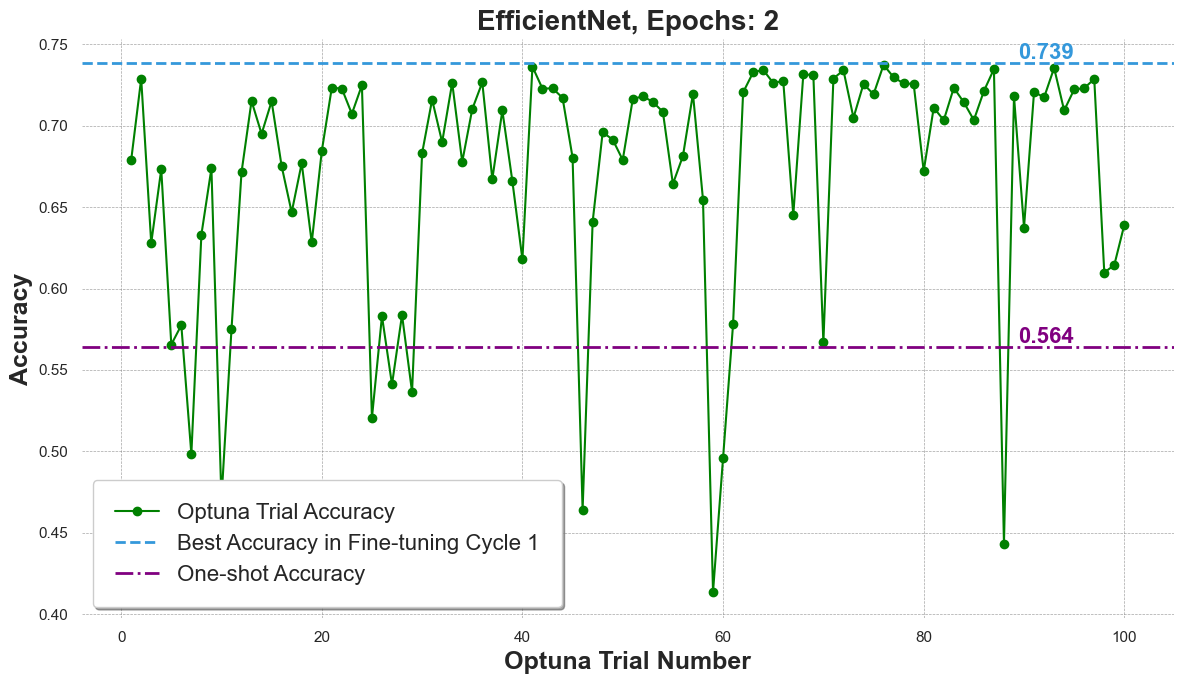}
            % \captionof{figure}{Image 4}
        \end{minipage} \\

        % Line 6
        \begin{minipage}{0.45\textwidth}
            \centering
            \includegraphics[width=\linewidth]{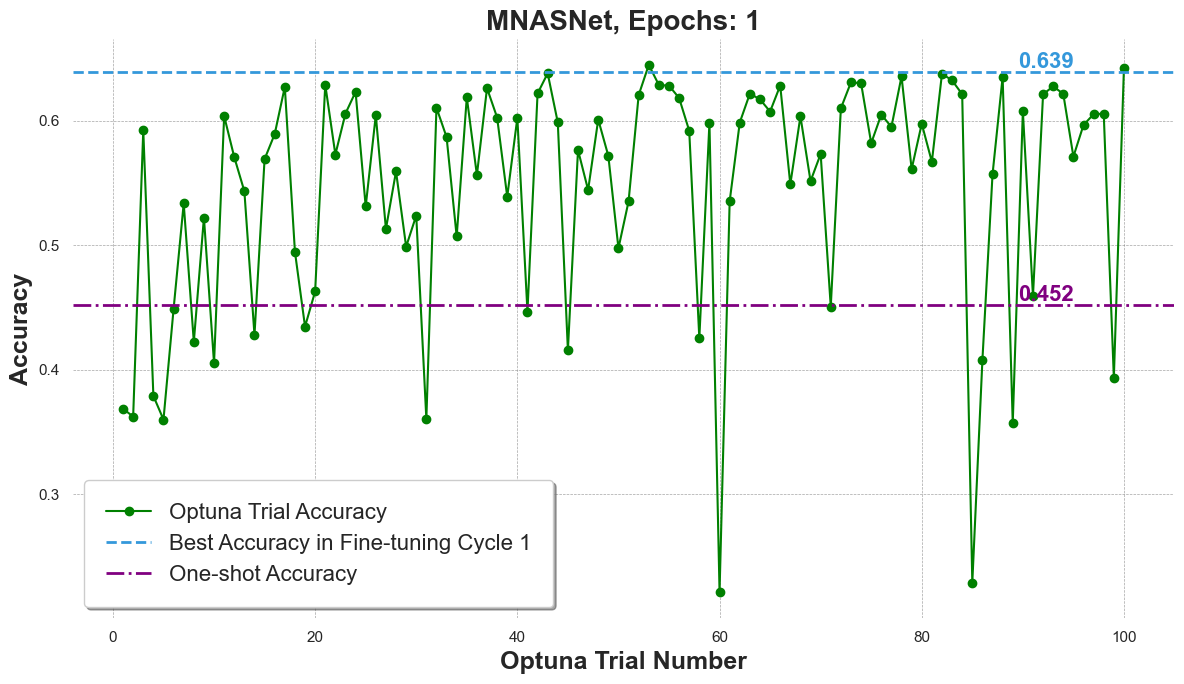}
            % \captionof{figure}{Image 3}
        \end{minipage} &
        \begin{minipage}{0.45\textwidth}
            \centering
            \includegraphics[width=\linewidth]{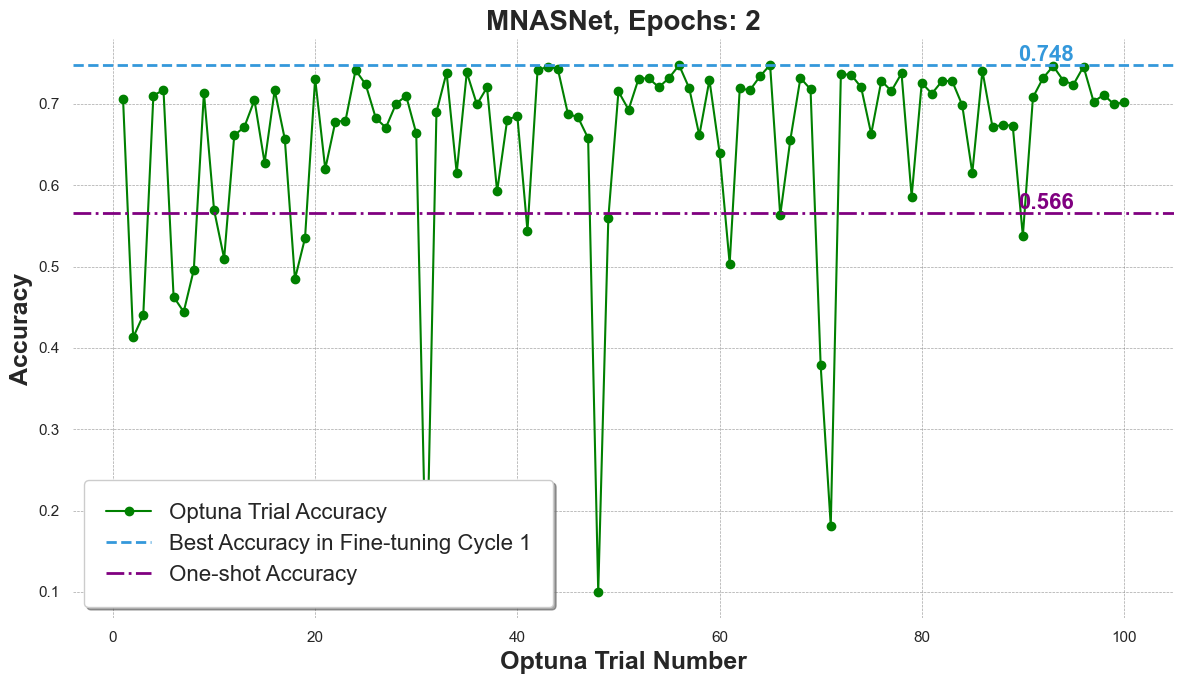}
            % \captionof{figure}{Image 4}
        \end{minipage} \\
        
    \end{tabular}
\caption{Part 1: Comparative analysis of accuracy performance for computer vision models evaluated over 100 trials using three distinct approaches: Optuna (green lines), fine-tuning with hyperparameters derived from Code Llama in Cycle 1 (blue lines), and one-shot predictions based on Code Llama (purple dashed lines). Each subplot represents a specific model and epoch configuration, illustrating variations in accuracy metrics and highlighting comparative trends.} 
\label{figure:10_8_models_Optuna_and_two_FT_lines_1}
\end{figure*}

\begin{figure*}[h!]
    \centering
    \setlength{\tabcolsep}{4pt}
    \renewcommand{\arraystretch}{10.0}
    \begin{tabular}{cc}

        % Line 7
        \begin{minipage}{0.45\textwidth}
            \centering
            \includegraphics[width=\linewidth]{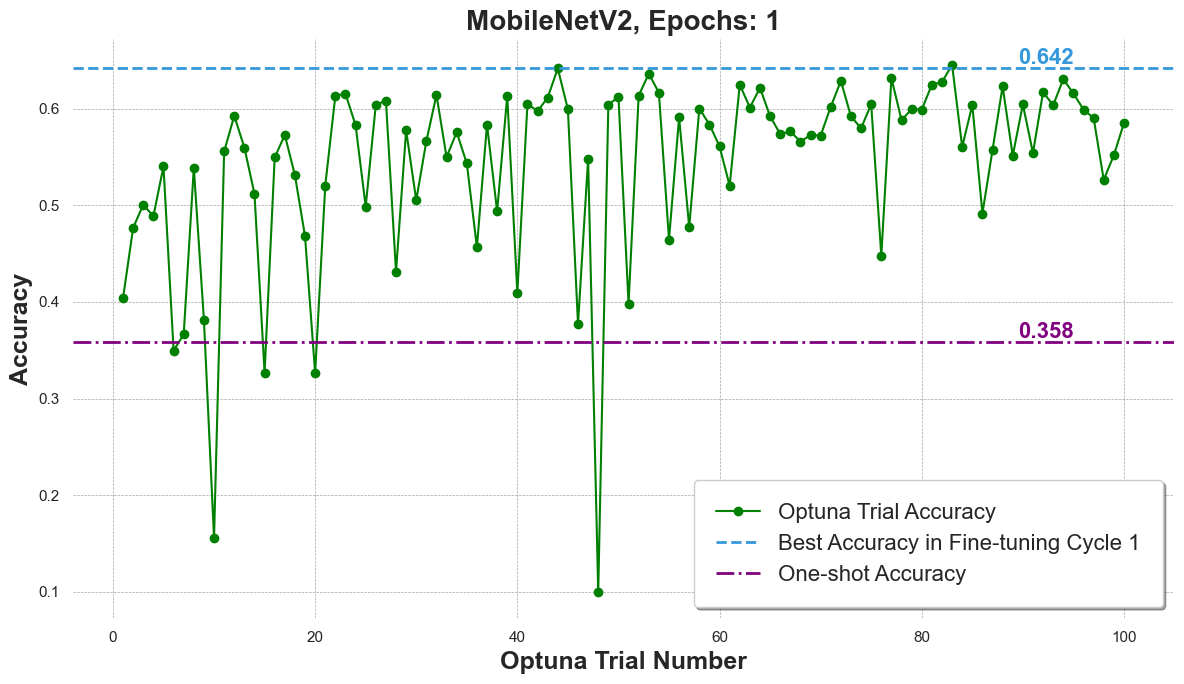}
            % \captionof{figure}{Image 3}
        \end{minipage} &
        \begin{minipage}{0.45\textwidth}
            \centering
            \includegraphics[width=\linewidth]{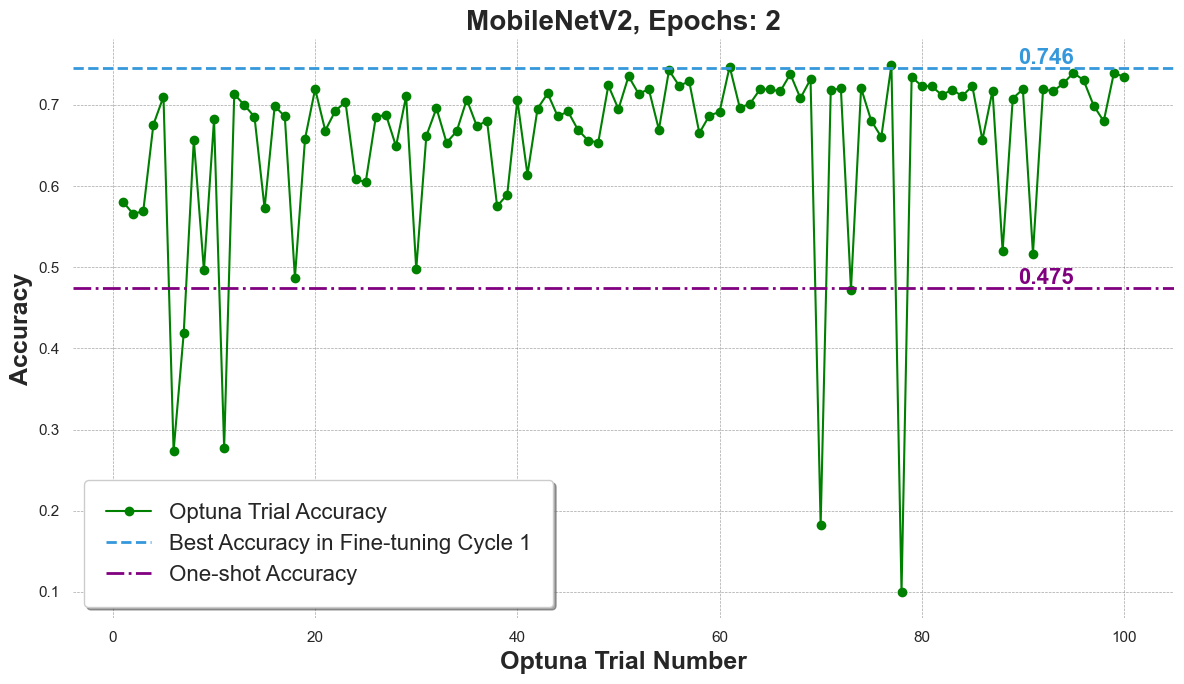}
            % \captionof{figure}{Image 4}
        \end{minipage} \\

        % Line 8
        \begin{minipage}{0.45\textwidth}
            \centering
            \includegraphics[width=\linewidth]{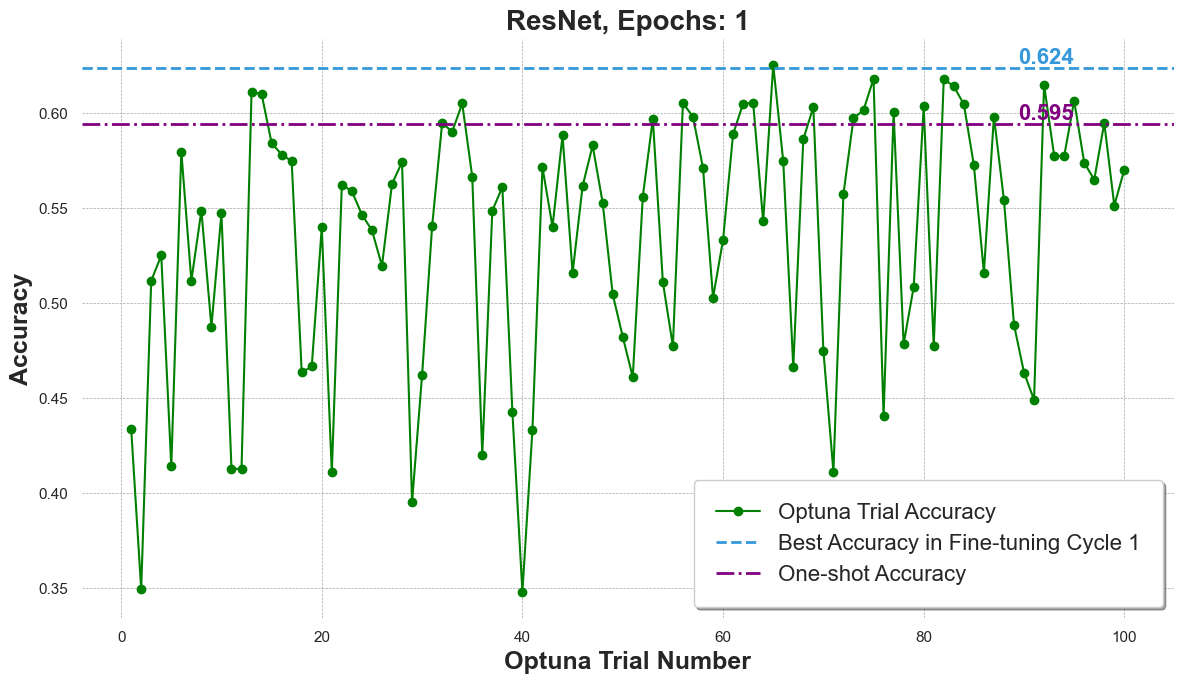}
            % \captionof{figure}{Image 3}
        \end{minipage} &
        \begin{minipage}{0.45\textwidth}
            \centering
            \includegraphics[width=\linewidth]{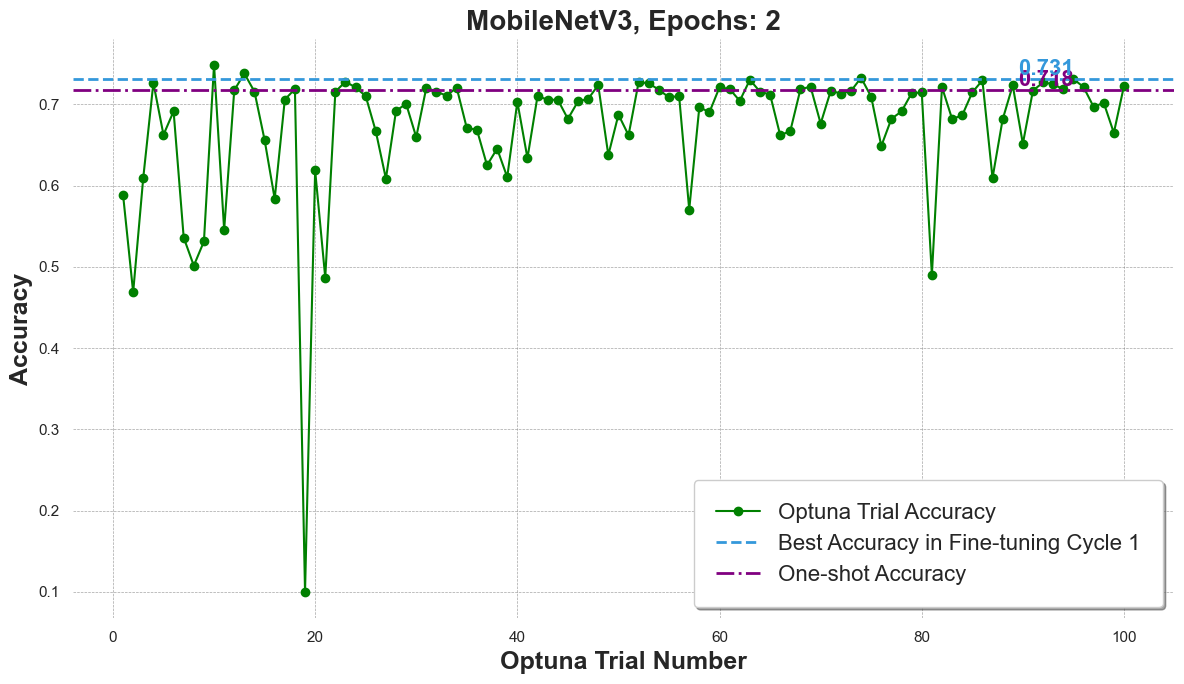}
            % \captionof{figure}{Image 4}
        \end{minipage} \\

        % Line 9
        \begin{minipage}{0.45\textwidth}
            \centering
            \includegraphics[width=\linewidth]{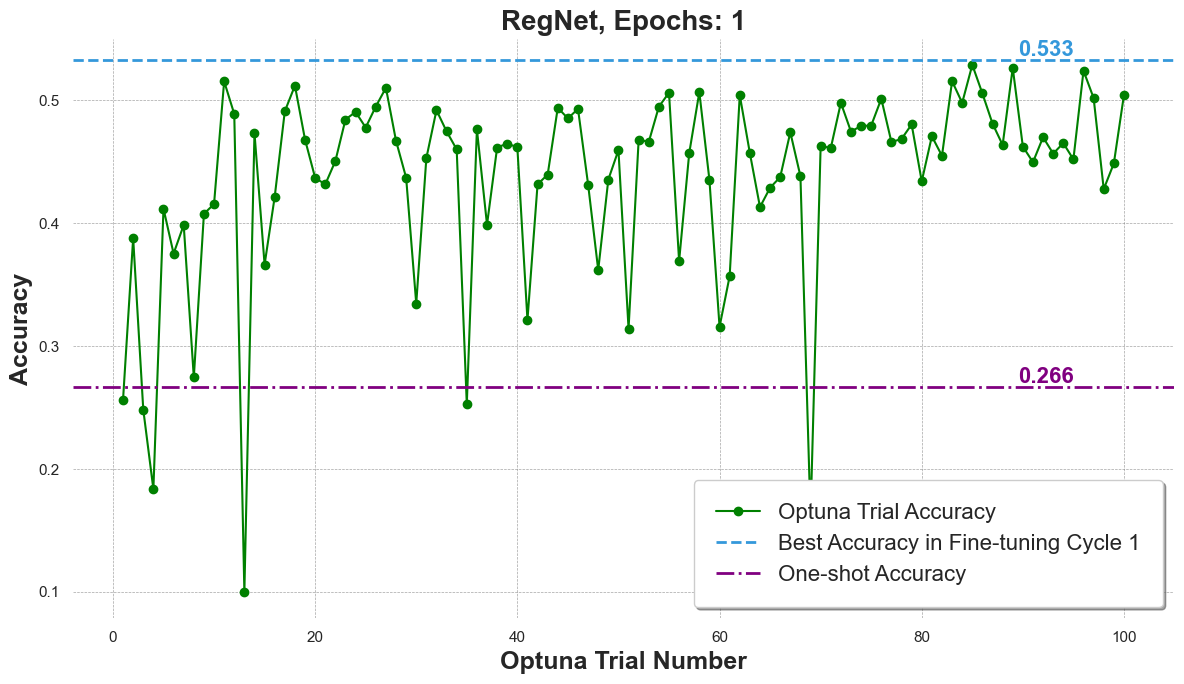}
            % \captionof{figure}{Image 3}
        \end{minipage} &
        \begin{minipage}{0.45\textwidth}
            \centering
            \includegraphics[width=\linewidth]{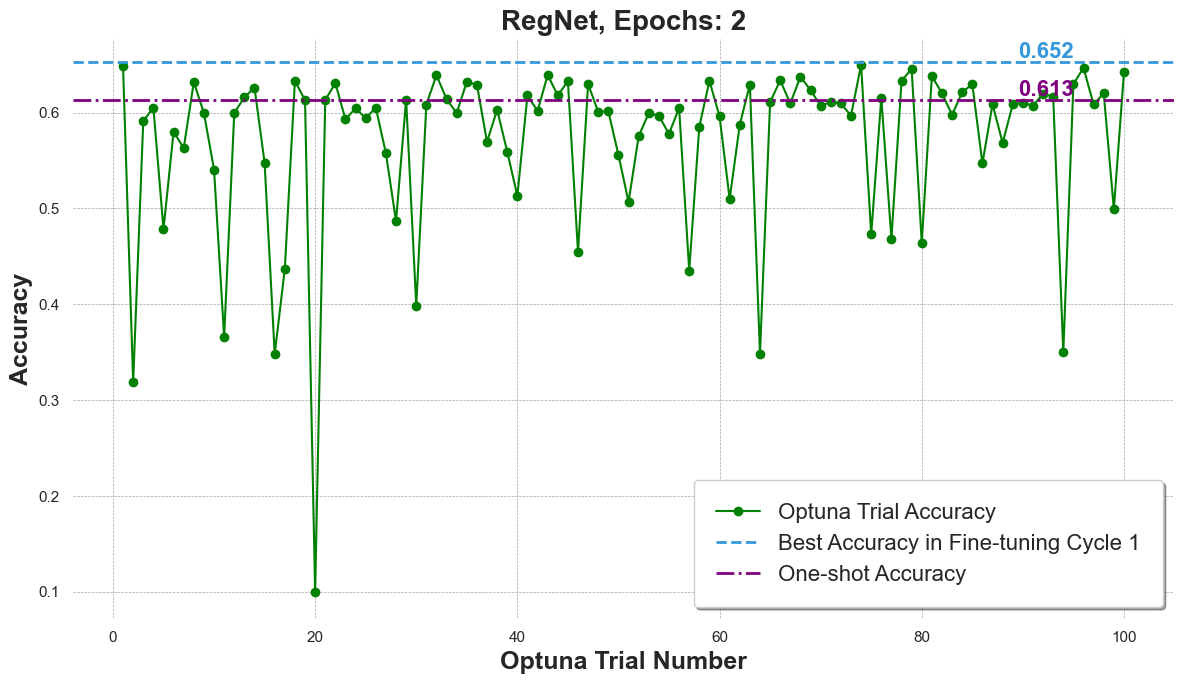}
            % \captionof{figure}{Image 4}
        \end{minipage} \\

        % Line 11
        \begin{minipage}{0.45\textwidth}
            \centering
            \includegraphics[width=\linewidth]{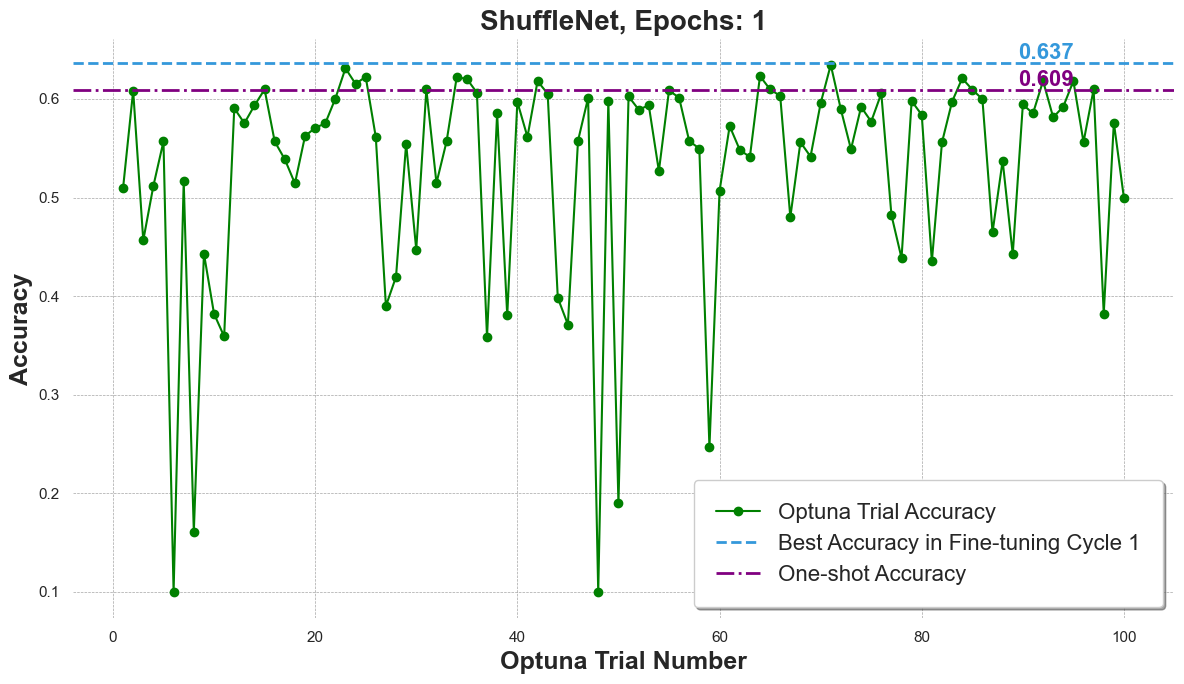}
            % \captionof{figure}{Image 3}
        \end{minipage} &
        \begin{minipage}{0.45\textwidth}
            \centering
            \includegraphics[width=\linewidth]{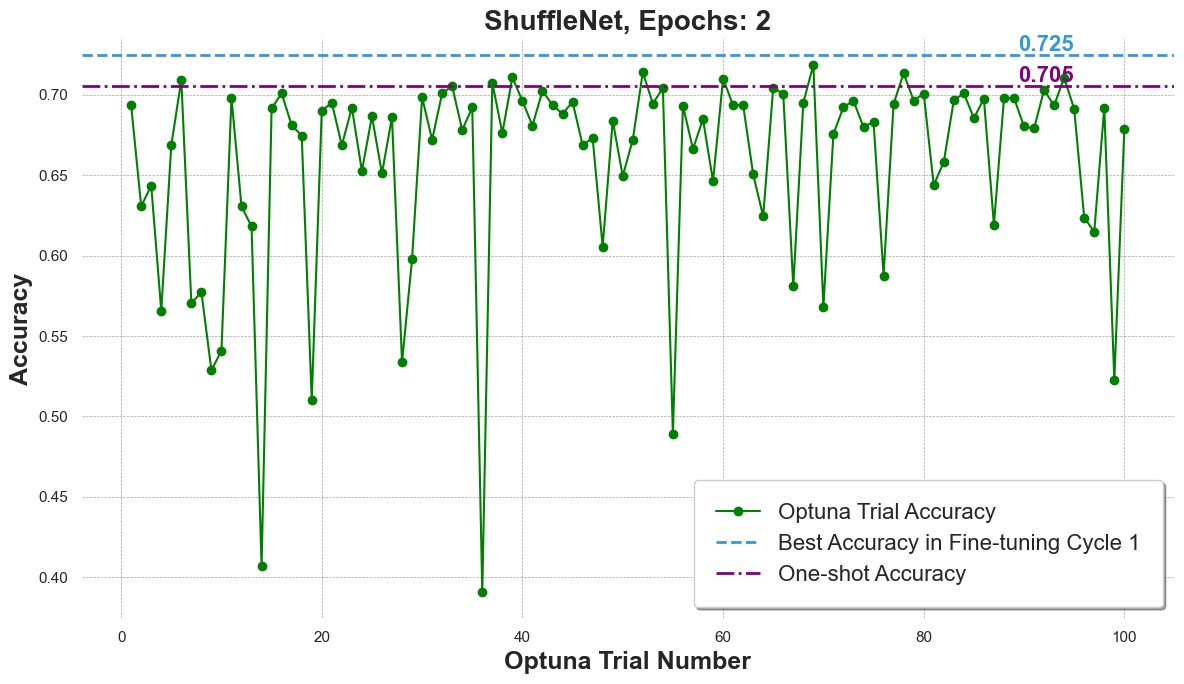}
            % \captionof{figure}{Image 4}
        \end{minipage} \\
    \end{tabular}
\caption{Part 2: Comparative analysis of the accuracy performance for the second set of computer vision models evaluated over 100 trials using three distinct approaches: Optuna (green lines), fine-tuning with hyperparameters derived from Code Llama in Cycle 1 (blue lines), and one-shot predictions based on Code Llama (purple dashed lines). Each subplot represents a specific model and epoch configuration, illustrating the variations in accuracy metrics and the relative performance across methodologies.} 
\label{figure:10_8_models_Optuna_and_two_FT_lines_2}
\end{figure*}

\begin{figure*}[h!]
    \centering
    \setlength{\tabcolsep}{4pt}
    \renewcommand{\arraystretch}{10.0}
    \begin{tabular}{cc}
    
        % Line 12
        \begin{minipage}{0.45\textwidth}
            \centering
            \includegraphics[width=\linewidth]{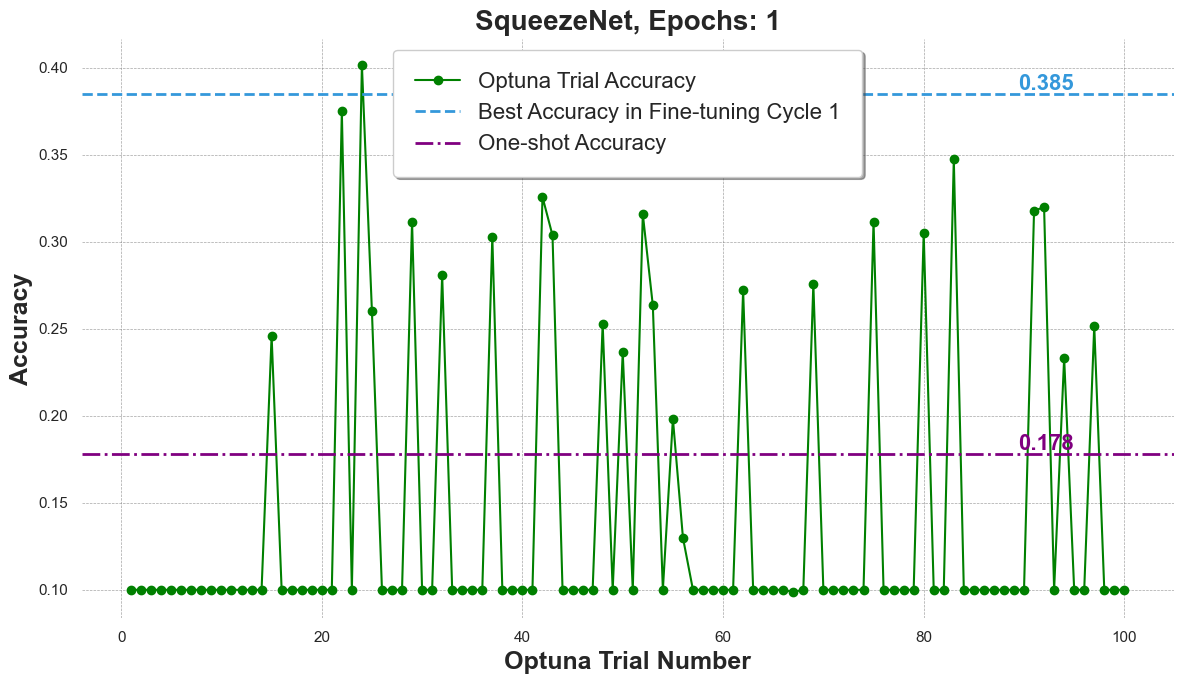}
            % \captionof{figure}{Image 3}
        \end{minipage} &
        \begin{minipage}{0.45\textwidth}
            \centering
            \includegraphics[width=\linewidth]{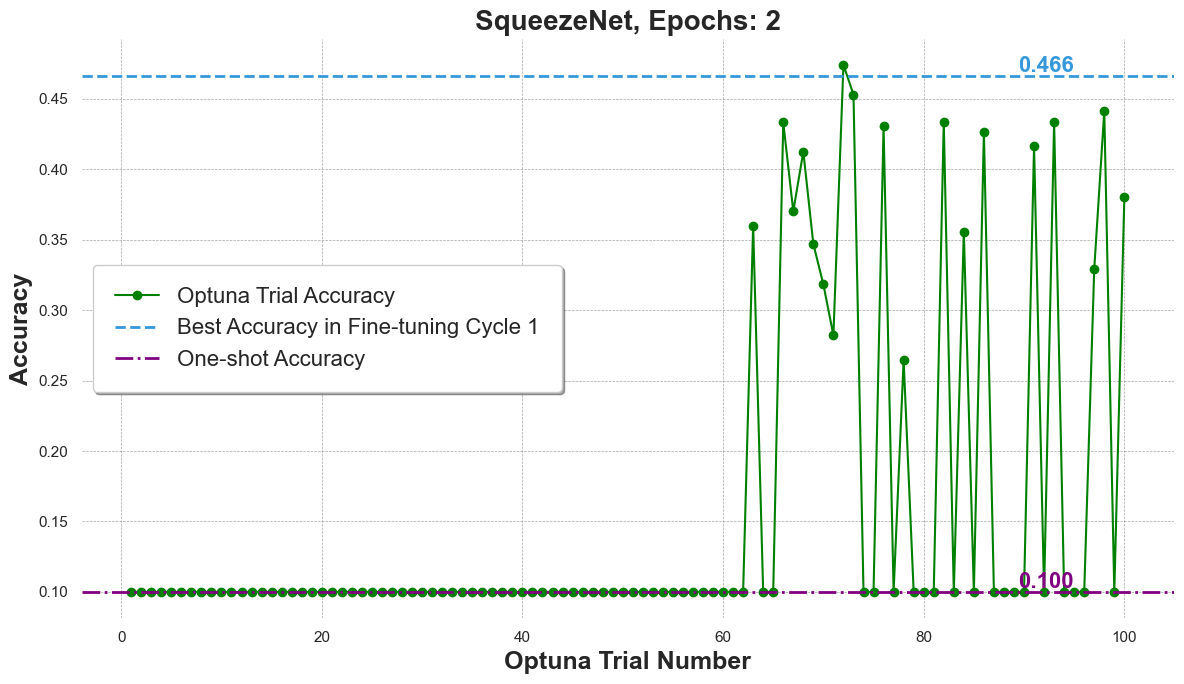}
            % \captionof{figure}{Image 4}
        \end{minipage} \\
        
        % Line 13
        \begin{minipage}{0.45\textwidth}
            \centering
            \includegraphics[width=\linewidth]{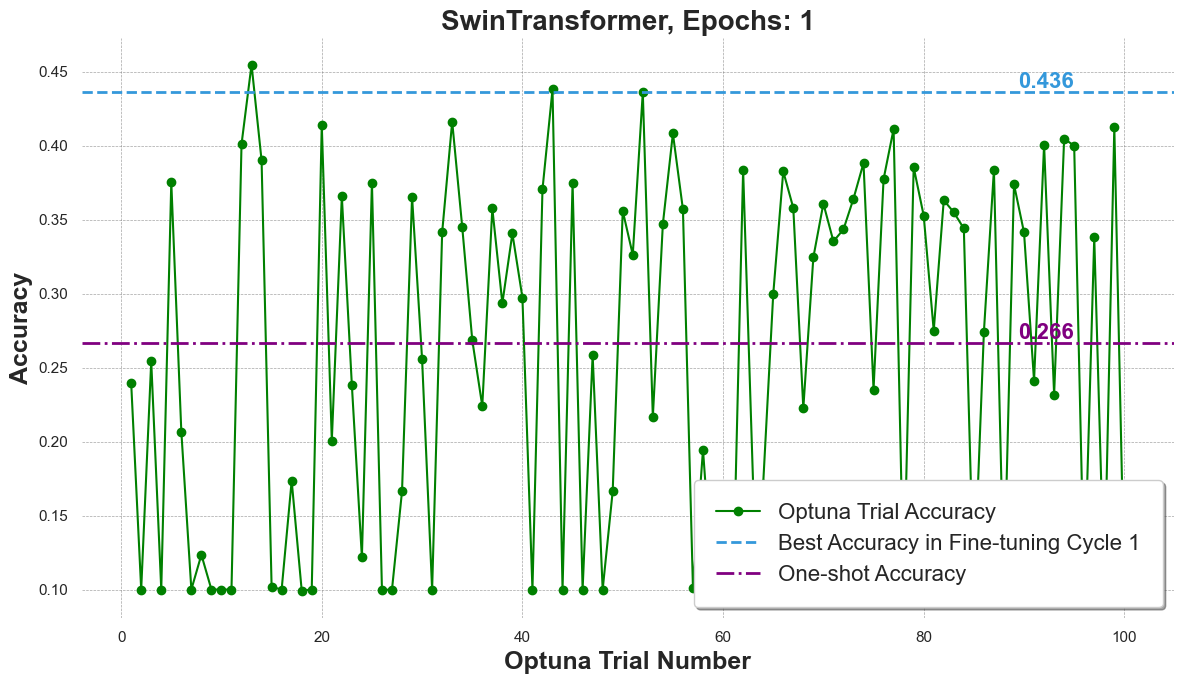}
            % \captionof{figure}{Image 3}
        \end{minipage} &
        \begin{minipage}{0.45\textwidth}
            \centering
            \includegraphics[width=\linewidth]{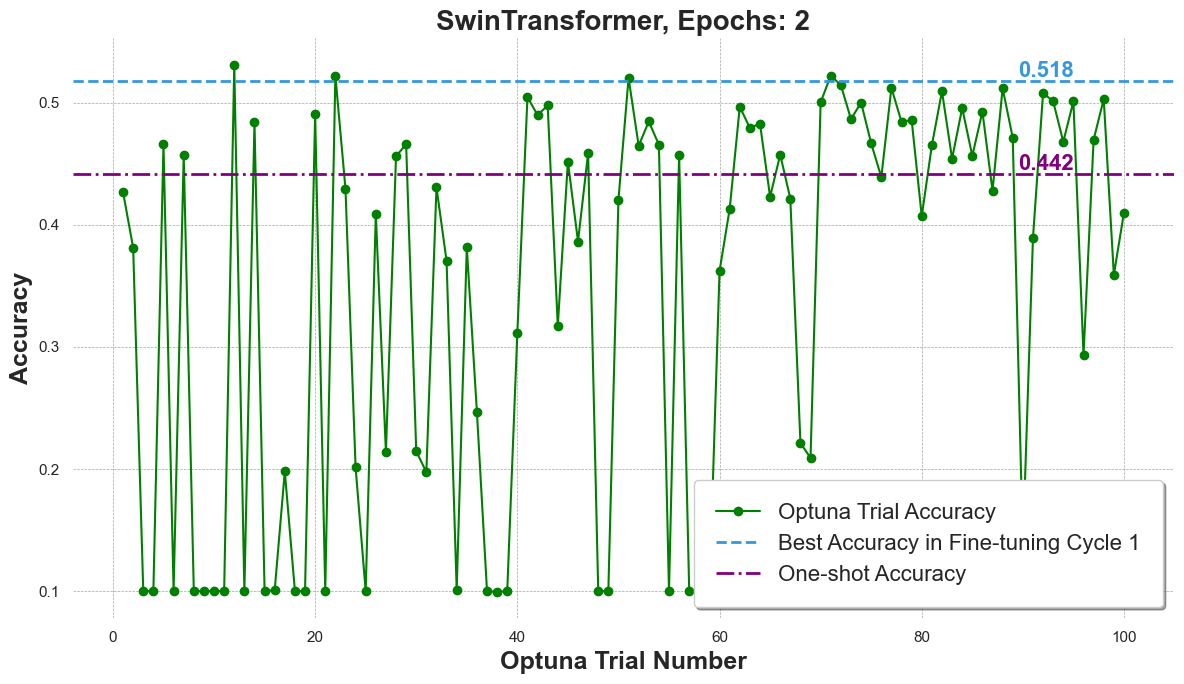}
            % \captionof{figure}{Image 4}
        \end{minipage} \\
        
        % Line 14
        \begin{minipage}{0.45\textwidth}
            \centering
            \includegraphics[width=\linewidth]{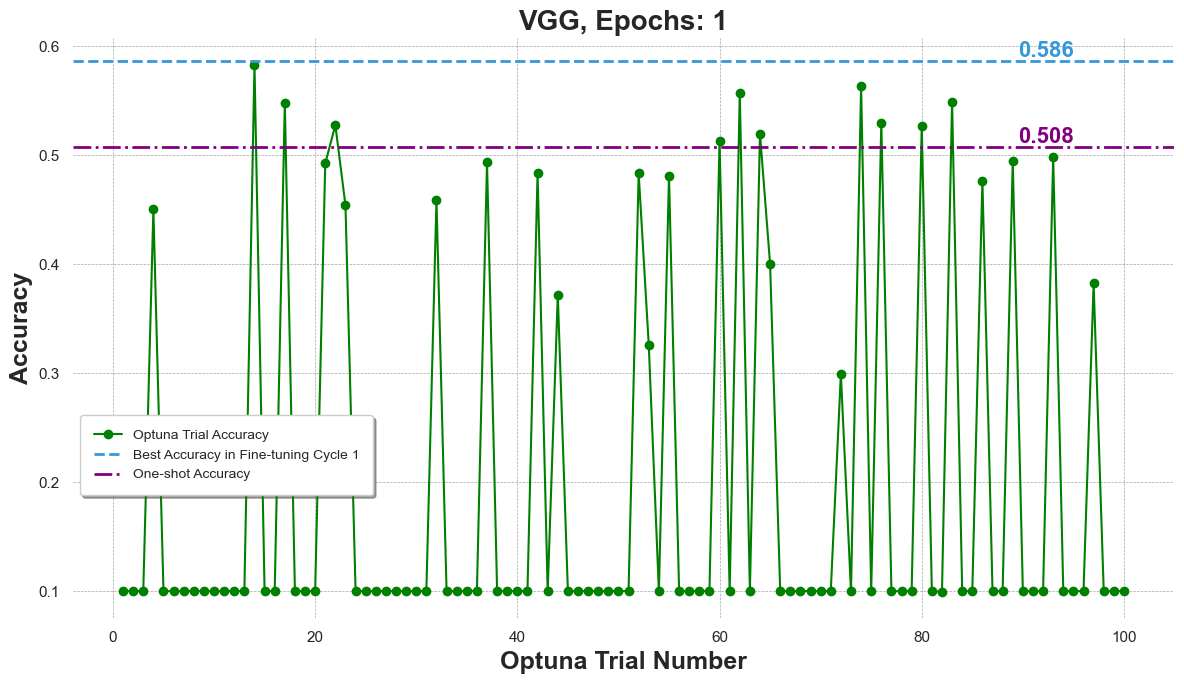}
            % \captionof{figure}{Image 27}
        \end{minipage} &
        \begin{minipage}{0.45\textwidth}
            \centering
            \includegraphics[width=\linewidth]{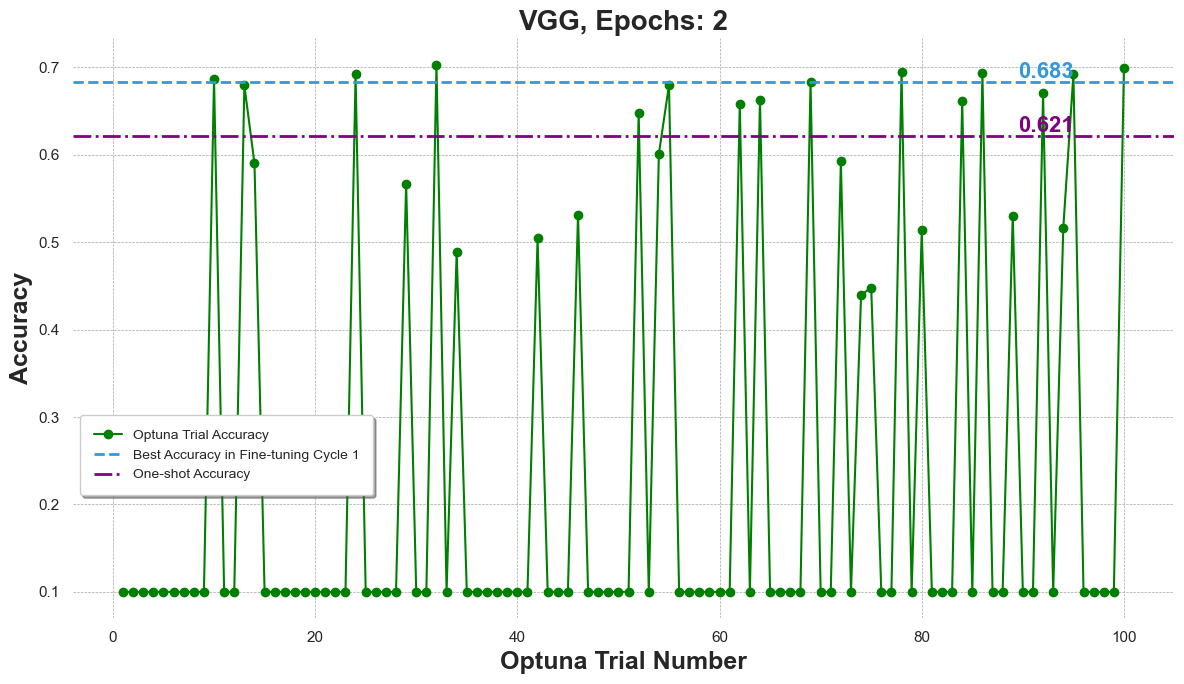}
            % \captionof{figure}{Image 28}
        \end{minipage} \\
    \end{tabular}
\caption{Part 3: Final analysis of accuracy dynamics for remaining computer vision models evaluated across 100 trials using three different approaches: Optuna (green lines), fine-tuning with hyperparameters obtained from Code Llama in Cycle 1 (blue lines), and one-shot predictions based on Code Llama (purple dashed lines). Each subplot corresponds to a specific model and epoch configuration, highlighting the variations in accuracy metrics.} 
\label{figure:10_8_models_Optuna_and_two_FT_lines_3}
\end{figure*}

The analysis of the accuracy performance of 14 computer vision models from the train dataset, each trained on 1 and 2 epochs (resulting in 28 graphs), excludes six graphs already covered in Section \ref{subsection:Training_Trends_and_Performance_Comparison} of this paper. The remaining 22 graphs, which are not presented in the main text, are illustrated in Figures \ref{figure:10_8_models_Optuna_and_two_FT_lines_1}-\ref{figure:10_8_models_Optuna_and_two_FT_lines_3}, showcasing key performance dynamics across models trained with Optuna over 100 trials, fine-tuned Code Llama, and one-shot predictions.

Fine-tuned Code Llama consistently delivers superior accuracy compared to Optuna’s results, as reflected by the blue line surpassing the scattered green points for almost all models. This highlights the fine-tuned model’s capability to optimize hyperparameters effectively. Pronounced improvements are observed in models like ResNet, ShuffleNet, and DenseNet, where the fine-tuning achieves significantly higher accuracy than Optuna’s best results.

Fine-tuned Code Llama also offers enhanced stability. Models like ConvNeXt, EfficientNet, and SwinTransformer, which show considerable variability in Optuna’s accuracy values, achieve more consistent performance with Code Llama. This consistency is particularly valuable in applications requiring dependable predictions across different conditions.

Increasing the number of epochs from one to two leads to expected accuracy improvements across all models. This trend is especially evident in models like EfficientNet and MobileNetV3, where additional training allows the parameters to refine further, resulting in higher accuracy.

One-shot predictions (purple line) demonstrate strong accuracy potential with minimal computational effort. In EfficientNet and MobileNetV3, the one-shot accuracy is nearly on par with the best results achieved through fine-tuning, emphasizing its viability as a quick and efficient method for generating robust predictions.

Certain models, such as SwinTransformer and SqueezeNet, show marked benefits from fine-tuning, overcoming the variability in their performance observed with Optuna. Code Llama fine-tuning reduces performance fluctuations and reliably enhances accuracy, even for architectures that are initially more challenging to optimize.

\end{document}